\documentclass[sigconf]{acmart}
\usepackage{enumitem}
\usepackage{multirow}
\usepackage{algorithm}
\usepackage{algorithmic}
\usepackage{color}
\usepackage{xcolor}
\usepackage{colortbl}
\usepackage[normalem]{ulem}
\useunder{\uline}{\ul}{}

\newtheorem{theorem}{Theorem}

\newtheorem{definition}{Definition}
\renewcommand{\algorithmicrequire}{ \textbf{Input:}}     %Use Input in the format of Algorithm
\renewcommand{\algorithmicensure}{ \textbf{Output:}}    %UseOutput in the format of Algorithm

\AtBeginDocument{%
  }

\setcopyright{acmlicensed}
\copyrightyear{2018}
\acmYear{2018}
\acmDOI{XXXXXXX.XXXXXXX}

\acmConference[Conference acronym 'XX]{Make sure to enter the correct
  conference title from your rights confirmation emai}{June 03--05,
  2018}{Woodstock, NY}
\acmISBN{978-1-4503-XXXX-X/18/06}

\begin{document}

\title{Toward Effective Digraph Representation Learning: A Magnetic Adaptive Propagation based Approach}

\author{Xunkai Li, Daohan Su, Zhengyu Wu, Guang Zeng, Hongchao Qin, Rong-Hua Li, Guoren Wang}

\renewcommand{\shortauthors}{Trovato et al.}

\begin{abstract}
    The $q$-parameterized magnetic Laplacian serves as the foundation of directed graph (digraph) convolution, enabling this kind of digraph neural network (MagDG) to encode node features and structural insights by complex-domain message passing.
    As a generalization of undirected methods, MagDG shows superior capability in modeling intricate web-scale topology.
    Despite the great success achieved by existing MagDGs, limitations still exist:
    (1) {Hand-crafted $q$}: 
    The performance of MagDGs depends on selecting an appropriate $q$-parameter to construct suitable graph propagation equations in the complex domain. 
    This parameter tuning, driven by downstream tasks, limits model flexibility and significantly increases manual effort.
    (2) {Coarse Message Passing}: 
    Most approaches treat all nodes with the same complex-domain propagation and aggregation rules, neglecting their unique digraph contexts.
    This oversight results in sub-optimal performance.
    To address the above issues, we propose two key techniques:
    (1) MAP is crafted to be a plug-and-play complex-domain propagation optimization strategy in the context of digraph learning, enabling seamless integration into any MagDG to improve predictions while enjoying high running efficiency.
    (2) MAP++ is a new digraph learning framework, further incorporating a learnable mechanism to achieve adaptively edge-wise propagation and node-wise aggregation in the complex domain for better performance.
    Extensive experiments on 12 datasets demonstrate that MAP enjoys flexibility for it can be incorporated with any MagDG, and scalability as it can deal with web-scale digraphs. MAP++ achieves SOTA predictive performance on 4 different downstream tasks.
\end{abstract}




\maketitle

\section{Introduction}
\label{sec: Introduction}
    As high-order structured data, the directed graph (digraph) offers a new perspective to model intricate web-scale information by capturing node relationships.
    Its exceptional representational capacity at the data level has driven advancements in graph mining at the model level, drawing significant attention in recent years~\cite{song2022gnn_survey4,platonov2023hete_gnn_survey4}.
    Notably, although existing undirected GNNs can achieve satisfactory performance, the loss of directed information undeniably limits their potential, especially when addressing topological heterophily challenges (i.e., whether connected nodes have similar features or same labels)~\cite{maekawa2023a2dug,dirgnn_rossi_2023,sun2023adpa}. 
    Therefore, researchers have increasingly focused on utilizing digraphs for modeling complex web scenarios, including recommendation~\cite{zhao2021ugrec_directed_app_recommendation1, Virinchi2023_directed_app_recommendation4} and social networks~\cite{bian2020_directed_app_social1, schweimer2022_directed_app_social2}. 
    Based on this, web mining problems can be translated into node-~\cite{tong2020digcn, zhang2021magnet, li2024lightdic}, link-~\cite{kollias2022digae, zhang2024dglp, ma2024dhgnn}, and graph-level~\cite{thost2021dagnn, liang2023hetdag, luo2024dagformer} tasks.

    To achieve effective digraph learning, a promising approach is $q$-parameterized magnetic Laplacian $\mathbf{L}_m$, which forms the foundation of digraph convolution from a spectral perspective to simultaneously encode node features and structural insights by message passing in the complex domain.
    Specifically, it is an adaptation of the standard Laplacian by incorporating complex-valued weights to account for the influence of a magnetic field on edges, which is particularly beneficial for investigating network properties when the edges are formulated as the asymmetry topology (e.g., digraphs)~\cite{chung2005spectral_graph_magnetic_laplacian1, chat2019spectral_graph_magnetic_laplacian2}.
    Notably, the weights of $\mathbf{L}_m$ denoted as $\exp\left(i \boldsymbol{\Theta}_{uv}^{(q)}\right)$, where $\boldsymbol{\Theta}_{uv}^{(q)}$ represents the magnetic potential or phase linked to the directed edge $e_{uv}$ and $q$ determines the strength of direction, reflecting the integration of the magnetic vector potential along the edge from node $u$ to $v$.
    Intuitively, it can also be viewed as the spatial phase angle between connected nodes in the complex domain, describing the direction and granularity of spatial message passing.

    Building upon this concept, digraph neural networks based on the $q$-parameterized magnetic Laplacian (MagDGs) implicitly execute eigen-decomposition during convolution~\cite{zhang2021magnet, zhang2021mgc, he2022msgnn, lin2023_framelet-magnet, zou2024_svd-magnet, li2024lightdic}. 
    This approach captures crucial structural insights (i.e., key properties of the digraph, such as connectivity) under the influence of the magnetic field, guiding optimal node encoding principles within the directed topology.
    Despite recent remarkable efforts in designing MagDGs, inherent limitations still exist:

    (1) \textit{Limited Understanding of $q$-parameterized Magnetic Laplacian in Digraph Learning}.
    Intuitively, $q$ determines the strength of direction for each edge in the digraph, manifested in the spatial phase angle between every connected node in the complex domain.
    For its direct impact on propagation and message (i.e., propagated results) aggregation, selecting an appropriate $q$ is crucial.
    However, related studies have primarily concentrated on spectral graph theory, providing guidance on $q$ selection from a strictly topological perspective and evaluating these principles in graph signal processing~\cite{furutani2020magLaplacian2}, community detection~\cite{fanuel2017q_magnetic3}, and clustering~\cite{fanuel2018magLaplacian1}.
    Despite their effectiveness, directly applying these methods in digraph learning is not suitable, as node profiles (i.e., node features and labels) are seldom considered in spectral graph theory and above applications. 
    In digraph learning, both node profiles and topology play equally crucial roles, and therefore, relying solely on topological measurements to define the $q$-parameterized magnetic Laplacian is insufficient and can mislead the message passing in the complex domain.
    To fill this gap, existing approaches treat $q$ as a hyperparameter, finely tuning it for different datasets and downstream tasks. 
    Although this strategy performs well in data-driven contexts, it often fails to thoroughly explore the optimal range of $q$, which increases manual cost, particularly in web-scale scenarios.

    \vspace{-0.05cm}
    \textbf{\textit{Solution}}:
    In Sec.~\ref{sec: Empirical Investigation} and Sec.~\ref{sec: Theoretical Analysis}, we conduct a comprehensive empirical study and theoretical analysis from topological and feature perspectives to explore the key insights behind the $q$-parameterized magnetic Laplacian in the context of digraph learning.

    (2) \textit{Lack of Fine-grained Message Passing in the Complex Domain}.
    Most existing methods directly utilize identical $q$ to achieve coarse-grained graph propagation in the complex domain. 
    This strategy assigns the same spatial phase angle to every directed edge, thereby employing the same propagation rules for all edges and neglecting their uniqueness.
    Furthermore, most approaches apply a simple averaging function during message aggregation after graph propagation. 
    This approach overlooks the varying contributions from different depths of structural insights encoded in the propagation, which are crucial for attaining optimal node representations.
    Obviously, this coarse-grained message passing in the complex domain leads to sub-optimal predictive performance.
    Meanwhile, real-world web mining applications with intricate directed topology heavily depend on the semantic contexts, which encompasses a comprehensive characterization based on their features and unique topology. 
    Hence, it is necessary to introduce a fine-grained message passing to capture such semantic context.

    \vspace{-0.05cm}
    \textbf{\textit{Solution}}.
    Motivated by the key insights obtained by Sec.~\ref{sec: Empirical Investigation}, we propose two pivotal techniques: 
    (i) MAP, a plug-and-play strategy seamlessly integrated with any existing MagDG, optimizes graph propagation in the complex domain through a weight-free angle-encoding strategy in the spatial phase, improving predictions while maintaining scalability. 
    (ii) MAP++, a new magnetic-based digraph learning framework, further quantifies the influence of node profiles and directed topology in the complex domain through a learnable strategy. 
    It achieves SOTA performance by flexible and adaptive edge-wise graph propagation and node-wise message aggregation.

    \textbf{Our contributions}.
    (1) \textit{\underline{New Perspective}}. 
    To the best of our knowledge, this paper is the first attempt to investigate the key insights of $q$-parameterized magnetic Laplacian in digraph learning. 
    We provide comprehensive empirical studies and highlight the integrated impact of node profiles and topology.
    (2) \textit{\underline{Plug-and-play Strategy}}. 
    We first propose MAP, which encodes spatial phase angles in a weight-free manner to tailor propagation rules for each node, seamlessly integrating with MagDGs to improve predictions. 
    (3) \textit{\underline{New Method}}. 
    To pursue superior performance, we propose MAP++, which utilizes learnable mechanisms to further optimize complex domain message passing, achieving edge-wise propagation and node-wise aggregation.
    (4) \textit{\underline{SOTA Performance}}. 
    Evaluations on 12 datasets, including large-scale ogbn-papers100M, prove that MAP has a substantial positive impact on prevalent methods (up to 4.81\% improvement) and MAP++ achieves the SOTA performance (up to 3.47\% higher).

\section{Preliminaries}
\label{sec: Preliminaries}

\subsection{Notations and Problem Formulation}
\label{sec: Notations and Problem Formulation}
    We consider a digraph $\mathcal{G}=(\mathcal{V}, \mathcal{E})$ with $|\mathcal{V}|=n$ nodes, $|\mathcal{E}|=m$ edges.
    Each node has a feature vector of size $f$ and a one-hot label of size $c$, the feature and label matrix are represented as $\mathbf{X}\in\mathbb{R}^{n\times f}$ and $\mathbf{Y}\in\mathbb{R}^{n\times c}$. 
    $\mathcal{G}$ can be described by an asymmetrical adjacency matrix $\mathbf{A}(u, v)$.  
    Downstream tasks include node-level and link-level.

\textbf{Node-level Classification.}
    Suppose $\mathcal{V}_l$ is the labeled set, the semi-supervised node classification paradigm aims to predict the labels for nodes in the unlabeled set $\mathcal{V}_u$ with the supervision of $\mathcal{V}_l$.

\textbf{Link-level Prediction.}
    (1) Existence: predict if $(u, v) \in \mathcal{E}$ exists in the edge sets;
    (2) Direction: predict the edge direction of pairs of nodes $u, v$ for which either $(u, v) \in \mathcal{E}$ or $(v, u) \in \mathcal{E}$;
    (3) Three-class link classification: classify an edge $(u, v) \in \mathcal{E},(v, u) \in \mathcal{E}$, or $(u, v),(v, u) \notin \mathcal{E}$.
    For convenience, we call it Link-C.

\textbf{Data-centric Plug-and-play MAP}: 
    This approach encodes spatial phase angles in a weight-free manner by considering the characteristics of digraph data from both topological and feature perspectives.
    It optimizes existing MagDGs by replacing their predefined rigid graph propagation equations (i.e., Hand-crafted $q$).

\textbf{Model-centric MAP++}: 
    Building on MAP, this method introduces additional learnable parameters to enable adaptive edge-wise graph propagation and node-wise message aggregation.
    The learnable modules from the above two perspectives can be selectively applied based on the computational capabilities, offering flexibility.

\subsection{Directed Graph Neural Networks}
\label{sec: Directed Graph Neural Networks}
\textbf{Prevalent Message Passing.}
    In undirected scenarios, prevalent approaches~\cite{hamilton2017graphsage, velivckovic2017gat, xu2018jknet, frasca2020sign, huang2020cands, li2024_atp} adhere to strict symmetric message passing. 
    This strategy entails the design of \textit{graph \textbf{Prop}agation} and the subsequent \textit{message \textbf{Agg}regation}, facilitating the establishment of relationships among a node and its neighbors.
    For the current node $u$, the $l$-th $\mathbf{W}$-parameterized {aggregator} is denoted as:
    \begin{equation}
    \label{eq: Prevalent undirected Message Passing}
        \begin{aligned}
        \mathbf{H}_u^{(l)} =\operatorname{Agg}\left( \mathbf{W}^{(l)},\operatorname{Prop}\left(\mathbf{H}_u^{(l-1)},\left\{\mathbf{H}_v^{(l-1)}, \forall v \in \mathcal{N}(u)\right\}\right)\right),
        \end{aligned}
    \end{equation}
    where $\mathbf{H}^{(0)}=\mathbf{X}$, $\mathcal{N}(u)$ denotes the one-hop neighbors of $u$. To obtain node embeddings in digraphs, it's crucial to consider the direction of edges. 
    Hence, the current node $u$ initially employs learnable weights separably for its out-neighbors $(u \rightarrow v)$ and in-neighbors $(v \rightarrow u)$ to obtain multi-level aggregated representations followed by the \textit{\textbf{Comb}ination} after directed message passing:
    \begin{equation}
    \begin{aligned}
        \label{eq: Prevalent directed Message Passing}
        \mathbf{H}_{u,\rightarrow}^{(l)}=&\operatorname{Agg}\left(\mathbf{W}_{\rightarrow}^{(l)},\operatorname{Prop}\left(\mathbf{H}_u^{(l-1)},\left\{\mathbf{H}_v^{(l-1)},\forall (u,v)\in\mathcal{E}\right\}\right)\right),\\
        \mathbf{H}_{u,\leftarrow}^{(l)}=&\operatorname{Agg}\left(\mathbf{W}_{\leftarrow}^{(l)},\operatorname{Prop}\left(\mathbf{H}_u^{(l-1)},\left\{\mathbf{H}_v^{(l-1)},\forall (v,u)\in\mathcal{E}\right\}\right)\right),\\
        &\;\;\mathbf{H}_u^{(l)} = \operatorname{Comb}\left(\mathbf{W}^{(l)}, \mathbf{H}_u^{(l-1)}, \mathbf{H}_{u,\rightarrow}^{(l)}, \mathbf{H}_{u,\leftarrow}^{(l)}\right).
        \end{aligned}
    \end{equation}
    Building upon this concept, DGCN~\cite{tong2020dgcn} and DiGCN~\cite{tong2020digcn} incorporate neighbor proximity to increase the receptive field (RF) of each node.
    DIMPA~\cite{he2022dimpa} increases the node RF by aggregating more neighbors during the graph propagation.
    NSTE~\cite{kollias2022nste} is motivated by the 1-WL graph isomorphism test to design the message aggregation.
    ADPA~\cite{sun2023adpa} explores appropriate directed patterns to conduct graph propagation.
    Despite their effectiveness, these methods inevitably introduce additional trainable weights and heavily rely on well-designed neural architectures that hinder their deployment.
    
\textbf{The $q$-parameterized magnetic Laplacian driven MagDGs.}
    To address these issues, recent studies employ the $q$-parameterized magnetic Laplacian to define complex-domain message passing, explicitly modeling both the presence and direction of edges through real and imaginary components. 
    Specifically, magnetic Laplacian is a complex-valued Hermitian matrix that encodes the asymmetric nature of a digraph via the $q$-parameterized complex part of its entries. 
    This introduces a complex phase, influenced by a magnetic field, to the edge weights, extending the conventional graph Laplacian into the complex domain to more effectively capture asymmetry.
    The above $q$-parameterized magnetic Laplacian is formally defined as:
    \begin{equation}
    \label{eq: magnetic Laplacian definition}
        \begin{aligned}
        &\;\;\;\;\;\;\;\;\mathbf{A}_m(u, v):=1/2\left(\mathbf{A}(u, v)+\mathbf{A}(v, u)\right),\\
        &\;\;\boldsymbol{\Theta}^{(q)}(u, v):=2 \pi q\left(\mathbf{A}(u, v)-\mathbf{A}(v, u)\right), q \geq 0,\\
        &\mathbf{L}^{(q)}_m:=\mathbf{D}_m-\mathbf{A}_m^{(q)}=\mathbf{D}_m-\mathbf{A}_m \odot \exp \left(i \boldsymbol{\Theta}^{(q)}\right),
        \end{aligned}
    \end{equation}
    where $\mathbf{D}_m$ is the degree matrix of $\mathbf{A}_m$, $q$ determines the strength of direction.
    The real part in $\mathbf{L}^{(q)}_m(u, v)$ indicates the presence and the imaginary part indicates the direction.
    Since we only consider unsigned digraphs, there exists $\cos\boldsymbol{\Theta}^{(q)}\ge 0$.
    Moreover, due to the periodicity of the $\sin\boldsymbol{\Theta}^{(q)}, \boldsymbol{\Theta}^{(q)}\in[-\pi/2,\pi/2]$,
    we have $q\in[0,1/4]$.
    When setting $q = 0$, directed information becomes negligible.
    For $q = 1/4$, we have $\mathbf{A}^{(q)}_m(u, v) =-\mathbf{A}^{(q)}_m(v, u)$ whenever there is an edge from $u$ to $v$ only.
    Based on this, we can formally define the magnetic graph operator (MGO) with self-loop ($\widetilde{\mathbf{A}}_m = \mathbf{A}_m+\mathbf{I}$) to form the foundation of digraph convolution as follows:
    \begin{equation}
    \label{eq: magnetic graph operator}
        \begin{aligned}
    \text{MGO}:=\hat{\mathbf{A}}_m=\left(\widetilde{\mathbf{D}}_m^{-1/2}\widetilde{\mathbf{A}}_m\widetilde{\mathbf{D}}_m^{-1/2} \odot \exp \left(i \boldsymbol{\Theta}^{(q)}\right)\right).
        \end{aligned}
    \end{equation}
    This MGO enables graph propagation in the complex domain, elegantly encoding deep structural insights concealed in digraphs with asymmetric topology.
    Subsequently, we can instantiate the trainable message aggregation based on the propagated results.
    The above $\mathbf{W}_{\mathbb{C}}$-parameterized \textit{\textbf{complex}}-domain message passing (proposed by MagNet~\cite{zhang2021magnet}) can be formally defined as:
    \begin{equation}
    \begin{aligned}
        \label{eq: MagDGs complex domain Message Passing}
        &\mathbb{C}_u^{(l-1)} = \operatorname{Complex}\left(\mathbf{H}_u^{(l-1)}\right):= \left\{\operatorname{Real}\left(\mathbf{H}_u^{(l-1)}\right), \operatorname{Imag}\left(\mathbf{H}_u^{(l-1)}\right)\right\},\\
        &\mathbb{C}_u^{(l)}=\operatorname{Agg}\left(\mathbf{W}_{\mathbb{C}}^{(l)},\operatorname{Prop}\left(\mathbb{C}_u^{(l-1)},\left\{\mathbb{C}_v^{(l-1)},\forall (u,v),(v,u)\in\mathcal{E}\right\}\right)\right).
        \end{aligned}
    \end{equation}
    Based on this foundation, MSGNN~\cite{he2022msgnn} extends this complex domain pipeline to directed signed graphs by varying the range of $q$. 
    MGC~\cite{zhang2021mgc} adopts a truncated version of PageRank named Linear-Rank to construct a filter bank to improve the graph propagation.
    Framelet-Mag~\cite{lin2023_framelet-magnet} employs Framelet-based filtering to decompose the magnetic Laplacian into components of different scales and frequencies for better predictive performance. 
    LightDiC~\cite{li2024lightdic} optimizes the MagDG framework by decoupling graph propagation and message aggregation for scalability in large-scale scenarios.

\begin{figure*}[t]   %注意，这里设置是关键
	\centering
    \setlength{\abovecaptionskip}{0.3cm}
	\includegraphics[width=\linewidth,scale=1.00]{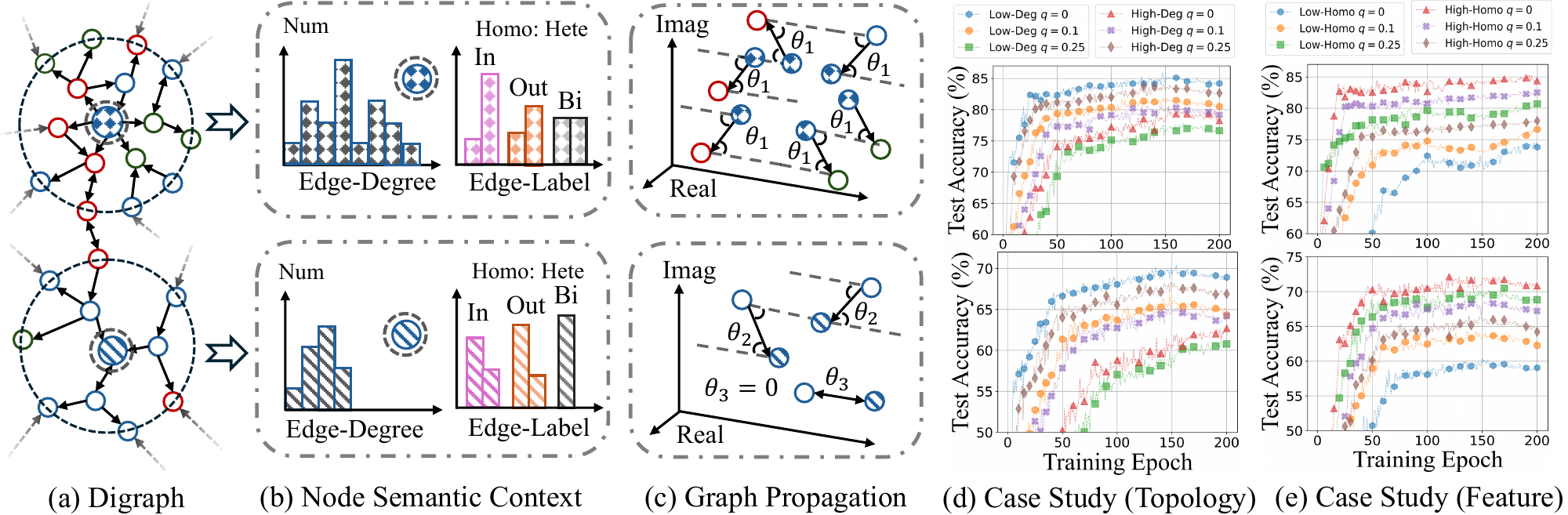}
	%[]里面的参数自己可根据需要调整
 \vspace{-0.5cm}
	\caption{
        (Left a,b,c) The illustration of different semantic contexts for two current nodes within two-hop neighbors, where colors denote label classes.
        (Right, d,e) The empirical study on directed CoraML (3.3k Nodes at Upper tables) and arXiv (169k Nodes at Lower tables) with different scales.}
	\label{fig: empirical studies}
 \vspace{-0.2cm}
\end{figure*}
\section{Empirical Investigation}
\label{sec: Empirical Investigation}
    As mentioned in Sec.~\ref{sec: Introduction} and Sec.~\ref{sec: Directed Graph Neural Networks}, despite the remarkable efforts of existing MagDGs in improving complex-domain graph propagation, two limitations still exist.
    To address them, we provide comprehensive empirical analysis in terms of:
    (1) \textbf{Illustrations}:
    We clarify the node semantic context driven by directed topology and visualize the naive graph propagation in existing MagDGs.
    Specifically, we choose two central nodes from Fig.~\ref{fig: empirical studies} (a) and perform statistical analysis from topological and feature perspectives in Fig.~\ref{fig: empirical studies} (b), where Edge-Degree denotes the sum of node degrees linked by two-hop edges of the current node and Edge-Label is the proportion of connected nodes with same or different labels (i.e., homophily and heterophily~\cite{ma2021hete_gnn_survey1, luan2022hete_gnn_survey2, zheng2022hete_gnn_survey3}) in different directions (i.e., incoming, outgoing, and bidirected edges). 
    Based on central nodes, we provide a visualization of the complex domain message passing in Fig.~\ref{fig: empirical studies} (c) highlighted by spatial phase angles (i.e., $\theta_1=\theta_2$).
    Notably, we select two representative digraph datasets of different scales for comprehensive comparison. 
    Compared to toy-sized Cora, large-scale arXiv better reflect the scalability challenges encountered in web-scale graph mining and the complexities of directed topology.
    (2) \textbf{Case Studies}:
    In Fig.~\ref{fig: empirical studies} (d) and (e), we use various magnetic parameters $q$ combined with 3-layer LightDiC~\cite{li2024lightdic} to evaluate the node performance with different semantic contexts across these two datasets.
    Similar to (1), we utilize node degrees and homophily to collectively support the semantic context.
    Specifically, in CoraML and arXiv, we classify nodes with degrees less than or equal to 3 and 5 as Low-Deg and other nodes as High-Deg, where Low-Deg at the digraph's periphery with fewer connections and High-Deg located at the center of densely connected communities.
    Meanwhile, for both datasets, we identify nodes with homophily less than and greater than 0.5 as Low-Homo and High-Homo, where node homophily~\cite{pei2020geomgcn} quantifies the similarity between the labels of the current node and its neighbors based on features, with higher values suggesting a higher probability of sharing the same label.

    \textbf{Observation 1:}
    \textit{Due to the intricate directed topology and feature correlation, nodes within the same digraph and RF may exhibit significantly diverse semantics.}
    As shown in Fig.~\ref{fig: empirical studies} (a,b), nodes within the same digraph and two-hop RF exhibit significant statistical disparities from both topological and feature perspectives, highlighting distinct contexts. 
    Intuitively, applying the same propagation rules to all nodes in digraphs inevitably results in high-bias performance.
    
    \textbf{Observation 2:}
    \textit{The predefined rigid edge-wise $q$ exacerbates the coarse-grained graph propagation above in the complex domain, further amplifying the adverse effects of overlooking the uniqueness of nodes.}
    Existing MagDGs adopt the same $q$ for all edges and assign identical phase angles to all node pairs in Fig.~\ref{fig: empirical studies} (c). 
    Given the semantic differences between the node and its neighbors with the complex computation between real and imaginary components, fine-grained propagation is necessary.

    \textbf{Key insight 1:}
    \textit{From the topological perspective, High-Deg poses greater prediction challenges than Low-Deg. 
    Fortunately, higher $q$ emphasizes direction, aiding High-Deg in discerning intricate neighborhoods.}
    In Fig. \ref{fig: empirical studies} (d), $q=0$ and $q=0.25$ respectively yield optimal performance for Low-Deg and High-Deg, as indicated by the blue and brown curves across two datasets.
    A deeper analysis can be pursued by investigating the trade-off of undirected ($\cos$) information and directed ($\sin$) information in graph propagation as described in Eq.~(\ref{eq: magnetic Laplacian definition}).
    For Low-Deg, $q=0$ results in a straight acquisition of knowledge from neighbors without additional directed information, thereby mitigating potential feature confusion issues arising from fewer neighbors.
    As for High-Deg, $q=0.25$ enables fine-grained discrimination of massive neighbors based on edge direction. 
    This facilitates the discovery of neighborhood knowledge that favors the current node for accurate predictions.

    \textbf{Key insight 2:}
    \textit{From the feature perspective, Low-Homo poses tougher prediction challenges than High-Homo. 
    Fortunately, higher $q$ facilitates Low-Homo for fine-grained propagation by emphasizing edge direction.}
    As depicted in Fig. \ref{fig: empirical studies} (e), we observe that $q=0$ and $q=0.25$ respectively result in optimal performance for High-Homo and Low-Homo, as indicated by the red and green curves.
    Notably, higher $q$ are particularly emphasized for discerning edge direction, especially in the context of the intricate directed topology of large-scale arXiv, depicted by the brown and green curves.
    For Low-Homo, $q=0.25$ enables effective differentiation between similar and dissimilar neighborhoods, thereby preventing the loss of node uniqueness due to naive propagation.
    As for High-Homo, which exhibits similarity among neighborhoods, $q=0$ achieves a straightforward yet effective approach to propagation, mitigating knowledge dilution introduced by additional directed information.

\section{Magnetic Adaptive Propagation}
\label{sec: Magnetic Adaptive Propagation}
    Motivated by the above key insights, in this paper, we propose two technologies: MAP and MAP++, offering a plug-and-play solution for existing MagDGs and a new MagDG framework, respectively.
    The core of our methods is the thorough integration of directed topology and node features, aimed at circulating the most appropriate magnetic field potential to directed edges. 
    In other words, we strive to ensure the quality of complex domain message passing by adaptive edge-wise graph propagation and node-wise message aggregation.
    Specifically, MAP first identifies the topological context of directed edges by quantifying the comprehensive centrality of start and end nodes, highlighting the direction of frequently activated edges (motivated by \textbf{Key insight 1}). 
    Subsequently, MAP quantifies the correlation between connected nodes in a weight-free manner throughout the edge projection in the complex plane.
    This process highlights the direction of edges linked by dissimilar nodes (motivated by \textbf{Key insight 2}).
    Building upon this foundation, MAP++ further introduces a learnable mechanism to achieve adaptive spatial phase angle encoding and weighted message aggregation to improve performance.
    The complete algorithm description and complexity analysis can be found in Appendix~\ref{appendix: MAP and MAP++ Algorithm and Complexity Analysis}.

\subsection{Topology-related Uncertainty Encoding}
\label{sec: Topology-related Uncertainty Encoding}
    Drawing from the empirical study, we conclude that frequently activated directed edges generate intricate information flows that compromise the uniqueness of node representations. 
    Based on this, we provide a more generalized and thorough perspective: 
    \textit{these intricate information flows driven by frequently activated directed edges introduce additional topological uncertainty to node representations, significantly disturbing their prediction, as evidenced by Fig.~\ref{fig: empirical studies} (d). }
    
    As {Key insight 1} highlighted, the directed information introduced by increased $q$ can be construed as supplementary encoding of topological uncertainty, thereby regulating graph propagation to avoid node confusion.
    In other words, this directed information enhances the capacity to discern complex information flows, enabling fine-grained graph propagation, and thereby improving node discrimination.
    Consequently, we aim to understand this topology-related uncertainty.
    It first identifies frequently activated directed edges through connected nodes and then applies fine-grained encoding to their magnetic field potentials for personalized propagation.

    In a highly connected digraph, nodes frequently interact with their neighbors. 
    By employing random walks~\cite{pearson1905random_walk}, we can capture these interactions and introduce Shannon entropy to measure node centrality~\cite{li2016structural_entropy_toit_16} from a global perspective. 
    Meanwhile, by adopting cluster connectivity, we can further offer a description of node centrality from a local perspective, which closely correlates with neighbor connectivity. 
    The above processes are defined as:
    \begin{equation}
    \label{eq: global centric and local centric for frequently activated edges}
    \begin{aligned} 
    &\;\;\;\;\;\operatorname{Global}:= GC(v)= \frac{\tilde{d}_v^{\mathrm{in}}}{m} \log \frac{\tilde{d}_v^{\mathrm{in}}}{m}+\frac{\tilde{d}_v^{\mathrm{out}}}{m} \log \frac{\tilde{d}_v^{\mathrm{out}}}{m},\\
    &\operatorname{Local}:= LC(v)=\mathbf{m}_v/\left(\tilde{d}_v^{\mathrm{in}}\cdot\tilde{d}_v^{\mathrm{out}}\right),\mathbf{m}_v=\sum_u\left(\tilde{\mathbf{A}}^2\odot\tilde{\mathbf{A}}^\top\right)_{vu},
    \end{aligned}
    \end{equation}
    where $\tilde{d}^{\mathrm{in}}$ and $\tilde{d}^{\mathrm{out}}$ are the in and out-degrees in the digraph.
    $\mathbf{m}_v$ is the triple motifs of node $v$.
    Notably, in contrast to directed structural entropy defined by the previous work~\cite{li2016structural_entropy_toit_16}, we address the limitation of only walking in the forward direction by incorporating reverse walking. 
    This modification is motivated by the non-strongly connected nature of most digraphs, where the proportion of complete walk paths declines sharply. 
    This decline suggests that most walk sequences fail to capture sufficient information beyond the immediate neighborhood of the starting node. 
    Consequently, strictly adhering to edge directions in walks (forward-only) results in severe walk interruptions, which ultimately degrades the effectiveness of $GC(v)$.
    Furthermore, we add self-loops for sink nodes to obtain $\tilde{\mathbf{A}}$. 
    This prevents the scenario where the adjacency matrix might be a zero power and ensures that the sum of landing probabilities is 1.
    
    Obviously, if connected nodes $u$ and $v$ exhibit large $LC$ and $GC$, they are positioned at the core of the digraph and contribute to the frequently activated $e_{uv}$ during graph propagation, which introduces topological uncertainty to node representations. 
    Notably, we have noticed that some spectral graph theory studies provide guidance on selecting $q$ from a strictly topological perspective. 
    However, it is crucial to emphasize that these methods are not directly applied in node profile-driven classification tasks, thereby inherent limitations are present.
    For more experimental results and analysis, please refer to Sec.~\ref{sec: Ablation Study}.
    To break this limitation, we directly assign a larger $q$ for $e_{uv}$ from the topological perspective and combine the subsequent feature-oriented encoding, which is defined as:
\begin{equation}
    \label{eq: q topology encoding}
    \begin{aligned} 
    q_{uv}^{\operatorname{topo}} = \operatorname{Norm}\left(GC_{u+v}+LC_{u+v}\right),\operatorname{Norm}(\mathbf{x})=\tanh{\frac{\mathbf{x}}{\operatorname{mean}(\mathbf{x})}}.
    \end{aligned}
\end{equation}

\subsection{Feature-related Correlation Encoding}
\label{sec: Feature-related Correlation Encoding}
    At this point, we have achieved topology-related uncertainly encoding for frequently activated directed edges.
    However, node features equally play a pivotal role in digraph learning.
    Therefore, we aim to fully leverage the correlation of connected nodes to further fine-tune the magnetic field potentials on directed edges.
    Motivated by Key insight 2, we conclude the following principles:
    (1) A smaller $q$ for connected nodes with high feature similarity, disregarding directed information to mitigate knowledge dilution.
    (2) A larger $q$ for connected nodes with low feature similarity, emphasizing directed information to enhance knowledge discernibly. 
    These principles enable the current node to acquire more beneficial knowledge.

    According to Eq.~(\ref{eq: magnetic Laplacian definition}), the complex plane is established by the $q$-parameterized magnetic Laplacian.
    Each directed edge is depicted as a vector within this complex plane and its projection on the $x$-axis(\textit{real part}) is edge existence, while the projection on the $y$-axis(\textit{imaginary part}) is edge direction.
    For connected nodes $u$ and $v$, dissimilar features lead to a larger $q$ with a greater angle for $e_{uv}$, indicating shorter projection along the $x$-axis and longer projection along the $y$-axis. 
    This emphasizes $e_{uv}$ direction during graph propagation and aligns with the previously mentioned principles.
    Consequently, we can directly leverage the correlation of features between connected nodes to encode the magnetic field potential of the corresponding directed edge, where node embeddings $\mathbf{Z}$ are obtained from $\mathbf{W}$-parameterized backbone MagDG.
    The above process can be formally defined as:
\begin{equation}
    \label{eq: q feature encoding}
    \begin{aligned} 
    q_{uv}^{\operatorname{feat}} = \operatorname{Norm}\left(\arccos\left(\frac{\mathbf{Z}_u\cdot\mathbf{Z}_v}{\Vert\mathbf{Z}_u \Vert\times \Vert \mathbf{Z}_v\Vert}\right)\right), \operatorname{Norm}(\mathbf{x}) = \frac{2\mathbf{x}}{\pi}.
    \end{aligned}
\end{equation}

\subsection{MAP Framework}
\label{sec: MAP Framework}
    Now, we have achieved fine-grained magnetic field potential encoding for directed edges, considering both topological and feature perspectives. 
    This is reflected in the adaptive spatial phase angles of connected nodes in the complex domain.
    To pursue scalability, we reformulate the originally rigid $q$-parameterized magnetic Laplacian from Eq.~(\ref{eq: magnetic Laplacian definition}) in a weight-free manner to obtain the optimized graph propagation kernel $\star\text{MGO}$. 
    It can be formally defined as:
    \begin{equation}
    \label{eq: map framework}
    \begin{aligned} 
    \hat{\mathbf{A}}_m^\star\!= \!\left(\widetilde{\mathbf{D}}_m^{-1/2}\widetilde{\mathbf{A}}_m\widetilde{\mathbf{D}}_m^{-1/2} \!\odot\! \exp \left(i \boldsymbol{\Theta}^{(\star q)}\right)\right), \!\star q=q^0\!\odot\! q^{\operatorname{feat}}\!\odot\! q^{\operatorname{topo}}
    \end{aligned}
\end{equation}
    where $q^0=1/4$ is the initial magnetic field potential parameter. Since $q^{\operatorname{feat}}$ and $q^{\operatorname{topo}}$ lie within the range $[0,1]$, $q^0$ can be adaptively scaled, thereby eliminating the need for manual adjustment.

\subsection{MAP++ Framework}
\label{sec: MAP++ Framework}
    Despite the progress made by MAP, the weight-free method often encounters limited improvement. 
    Furthermore, most MagDGs directly stack linear layers to implement message passing, resulting in strict dependencies between the current and the previous layer.
    This coupled architecture can only support shallow MagDGs with limited RFs and toy-size datasets, as deeper ones would suffer from the over-smoothing problem, out-of-memory (OOM) error, and out-of-time (OOT) error, especially in web-scale sparse digraphs.
    To break the above limitations, we propose MAP++ as follows:

\textbf{Step 1: Edge-wise Graph Propagation.}
    Based on the MAP, we first utilize a lightweight neural architecture $\operatorname{Edge-Mag}(\cdot)$ parameterized by $\mathbf{W}_{edge}$ to further encode magnetic field potentials for each directed edge.
    In this strategy, we aim to enable iterative optimization through the training, which is formally defined as:
\begin{equation}
    \label{eq: map++ q topology encoding}
    \begin{aligned} 
    \star q\! =q^0\!\odot\!\operatorname{Edge-Mag}\left(\operatorname{Norm}\left(GC_{u+v}\odot q_{uv}^{\operatorname{feat}}\Vert LC_{u+v}\odot q_{uv}^{\operatorname{feat}}\right)\right),
    \end{aligned}
\end{equation}
    where $\odot$ denotes the element-wise matrix multiplication.
    Notably, this approach is only for small- and medium-scale datasets due to scalability.
    To increase the RF of nodes, we conduct $K$-step complex-domain graph propagation, correspondingly getting a list of propagated features (i.e., messages) under different steps as follows:
\begin{equation}
    \label{eq: map++ feature propagation}
    \begin{aligned} 
    \widetilde{\mathbf{X}}^{(K)}=\hat{\mathbf{A}}_m^{\star K}\widetilde{\mathbf{X}}^{(0)}\rightarrow[\widetilde{\mathbf{X}}^{(0)},\widetilde{\mathbf{X}}^{(1)},\dots,\widetilde{\mathbf{X}}^{(K)}],\widetilde{\mathbf{X}}^{(0)} = \mathbf{X}.
    \end{aligned}
\end{equation}
    Due to the learnable $\hat{\mathbf{A}}_m^{\star K}$, gradients flow towards propagated features.
    Thus far, we have achieved edge-wise graph propagation by integrating adaptive magnetic field potential during training.

\textbf{Step 2: Node-wise Message Aggregation.}
    Recent studies~\cite{frasca2020sign,sun2021sagn,zhang2021rod} have highlighted that the optimal RF varies for each node, influenced by the intricate semantic context. 
    This insight is especially critical for digraphs in the complex domain, where multi-level structural encoding in Eq.(\ref{eq: map++ feature propagation}) often provides valuable prompts within the coupling of real and imaginary components.
    Therefore, we advocate explicitly learning the importance and relevance of multi-granularity knowledge within different RF in a node-adaptive manner to boost predictions.
    This process can be defined as follows:
\begin{equation}
    \label{eq: map++ message aggregation}
    \begin{aligned} 
    &\;\;\mathbf{H} = \sum_{l=0}^K\mathbf{W}_{node}^{(l)}\widetilde{\mathbf{X}}^{(l)},\mathbf{W}_{node}^{(l)}=e^{\delta\left(\mathbf{E}^{(l)}\right)}/\sum_{i=0}^K e^{\delta\left(\mathbf{E}^{(i)}\right)},\\
    &\mathbf{E}^{(l)} = \operatorname{MLP}\left(\operatorname{Complex}\left(\widetilde{\mathbf{X}}^{(0)}\right)\Vert\dots\Vert\operatorname{Complex}\left(\widetilde{\mathbf{X}}^{(K)}\right)\right),
\end{aligned}
\end{equation}
    where $\delta$ is the non-linear activation function. 
    This mechanism is designed to construct a personalized multi-granularity representation fusion for each node, facilitating the weighted message aggregation. 
    As the training progresses, the MAP++ gradually accentuates the importance of neighborhood regions in the complex domain that contribute more significantly to the target nodes.

\section{Theoretical Analysis}
\label{sec: Theoretical Analysis}
    Now, we have achieved adaptive magnetic field potential modeling for directed edges.
    To further investigate the effectiveness of our approach and ensure theoretical interpretability, we build upon insights from related studies~\cite{singer2011graph_angular_synchronization1, he2023graph_angular_synchronization4} by extending the angular synchronization framework to graph attribute synchronization problem, which incorporates node features and directed topology.

\noindent\textbf{Graph Attribute Synchronization.} 
    The conventional angular synchronization problem aims to estimate a set of unknown angles $\theta_1, \ldots, \theta_n$ from $m$ noisy measurements of their pairwise offsets~\cite{singer2011graph_angular_synchronization1}. 
    The noise associated with these measurements is uniformly distributed over the interval $[0,2 \pi)$. 
    Based on this, we have:
\vspace{-0.1cm}
\begin{definition}
    In the graph $\mathcal{G}=\{\mathcal{V}, \mathcal{E}\}$, each node $u \in \mathcal{V}$ is associated with an angle $\theta_u$. 
    Given noisy measurements of angle offsets $\delta_{i j}$, the angular synchronization problem aims to estimate the angles $\theta_1, \ldots, \theta_n$. 
    The distribution of $\delta$ is divided into two categories: reliable (good) edges $\mathcal{E}_{\text {good }}$ and unreliable (bad) edges $\mathcal{E}_{\text {bad }}$
\begin{equation}
\begin{aligned}
&\delta_{ij}=\theta_i-\theta_j \;\;\;\; {\rm for} \; (i,j)\in\mathcal{E}_{good} \\
\delta_{ij}\sim U&niform\left([0,2\pi)\right) \;\;\;\; {\rm for} \; (i,j)\in\mathcal{E}_{bad}.   
\end{aligned} 
\end{equation}
\end{definition}
    Based on this, the adaptive phase matrix in MAP functions as a weighted adjacency matrix, reflecting the presence of edges and capturing the offsets, analogous to $\delta$. 
    By treating it as a noisy node feature offset matrix, we can generate attribute $w_u$ for each node $u$ based on the node features and directed topology and have:
\vspace{-0.1cm}
\begin{definition}
    The graph attribute synchronization problem aims to estimate a set of unknown attributes $w_1, \ldots, w_n$ based on their noisy adaptive complex-domain offsets $\boldsymbol\Theta^{\left(q^{\star}\right)}$, which are defined as:
\begin{equation}
w_u-w_v:=2\pi q_{uv}^\star\left(\mathbf{A}_{uv}-\mathbf{A}_{vu}\right).
\end{equation}
\end{definition}
    This formulation demonstrates how the attributes $w_u$ can be inferred from the topology and phase information by leveraging the feature-related relationships between nodes. 
    For the numerous zero values in the matrix $\boldsymbol\Theta^{\left(q^\star\right)}$, we treat them as noisy data.

\noindent
\textbf{Spectral Analysis in MAP.} 
    According to the related studies~\cite{singer2011graph_angular_synchronization1,cucuringu2012graph_angular_synchronization2,cucuringu2016graph_angular_synchronization3}, solving the above graph attribute synchronization problem typically involves constructing a Hermitian matrix. 
    We first investigate the MAP encoding process (see Sec.~\ref{sec: Topology-related Uncertainty Encoding}-\ref{sec: Feature-related Correlation Encoding}) and have:
\begin{theorem}
\label{theorem: 1}
    The adaptive phase matrix $\boldsymbol\Theta^{(q^\star)}$ encoding by MAP is skew-symmetric, and $\mathbf{H}$ is Hermitian, where $\mathbf{H}=\exp\left(i\boldsymbol\Theta^{(q^\star)}\right)$.
\end{theorem}
    Based on this, we define the optimization objective as:
\begin{equation}
    \hat w=\arg\max(w)\sum^n_{i,j=1}e^{-iw_i}\mathbf{H}_{ij}e^{iw_j}.
\end{equation}
    This formulation effectively captures complex-domain offsets from topology and feature perspectives.
    However, it remains a non-convex problem, making it difficult to solve in practice.
    Here, we introduce the relaxation: let $z_i=e^{i w_i}$ and impose the constraint $\sum_{i=1}^n\left|z_i\right|^2=n$. 
    This leads to the following optimization objective:
\begin{equation}
    \hat{z}=\arg \max(z)\; z^*\mathbf{H}z.
\end{equation}
    Obviously, the maximizer $z$ is given by $z=v_1$, where $v_1$ is the normalized top eigenvector satisfying $\mathbf{H}v_1=\lambda_1v_1$ and $||v_1||^2=n$, where $\lambda_1$ is the largest eigenvalue of $\mathbf{H}$.
    Thus, the estimated attributes can be defined as: $e^{i\hat{w}_i}=v_1(i)/|v_1(i)|$.

    Although the adaptive phase matrix contains noise, which may cause discrepancies in estimations, but these discrepancies decrease as the noise reduces.
    Notably, even with significant noise, the eigenvector method can effectively recover attributes given enough noise-free equations. 
    Furthermore, we demonstrate that if the adaptive phase matrix is devoid of noise, the estimated attributes correspond to true attributes.
    Based on this, we have the following theorems.
\begin{theorem}
\label{theorem: 2}
    The correlation between the estimated node attributes $v_1$ and the true attributes $z$ is positively correlated with the number of nodes and inversely proportional to the square of the noise rate.
\end{theorem}
\begin{theorem}
\label{theorem: 3}
    If $\boldsymbol\Theta$ is noise-free, $e^{i \hat{w}_i}=\frac{v_1(i)}{\left|v_1(i)\right|}$ represents the unique exact solution to the graph attribute synchronization problem.
\end{theorem}
    Until now, we have provided the generalization of MAP to the graph attribute synchronization, offering theoretical robustness to our approach.
    Notably, traditional methods often rely on spectral methods based on rigid topology analysis to assign fixed $q$ for each edge (see Appendix~\ref{appendix: Guidance for Selecting q in Spectral Graph Theory}), which limits the flexibility and adaptability of the synchronization process.
    In contrast, MAP enables personalized $q$ values for each edge, considering not only the direction but also encoding uncertainty and correlation. 
    In a nutshell, MAP significantly enhances optimization capabilities for attribute synchronization by offering a more nuanced approach to the assignment of $q$.
    For detailed proofs of the above theorems, please refer to Appendix~\ref{appendix:The Proof of Theorem1}-\ref{appendix:The Proof of Theorem3}. 
    Additionally, we acknowledge that the recently proposed GNNSync~\cite{he2023graph_angular_synchronization4} also provides a theoretical analysis from the perspective of graph attribute synchronization. 
    For a further discussion of our approach and GNNSync, please see Appendix~\ref{appendix: Our Approach and GNNSync}.

\section{Experiments}
\label{sec: Experiment}
%% 可视化实验（热力图）
%% 更多 undirected gnn baseline 
%% link-level 实验结果
    In this section, we aim to offer a comprehensive evaluation and address the following questions to verify the effectiveness of our proposed MAP and MAP++:
    \textbf{Q1}: As a hot-and-plug strategy, what is the impact of MAP on the existing MagDGs?
    \textbf{Q2}: How does MAP++ perform as a new digraph learning model?
    \textbf{Q3}: If MAP and MAP++ are effective, what contributes to their performance?
    \textbf{Q4}: What is the running efficiency of them?
    \textbf{Q5}: How robust is MAP and MAP++ when dealing with sparse scenarios?
    To maximize the usage for the constraint space, we will introduce datasets, baselines, and experiment settings in Appendix~\ref{appendix: Dataset Description}-\ref{appendix: Experiment Environment}.

\subsection{Performance Comparison}
\label{sec: Performance Comparison}
\textbf{A Hot-and-plug Optimization Module.}
    To answer \textbf{Q1}, we present the performance enhancement facilitated by MAP in Table~\ref{tab: map performance improvement} and Table~\ref{tab: map link performance improvement}. 
    We observe that MAP significantly benefits all methods.
    This is attributed to its adaptive encoding of magnetic field potentials for directed edges, thereby customizing propagation rules. 
    Notably, due to the different numerical ranges of the metrics, the improvements at the node level are more pronounced.
    Meanwhile, the coupling architectures and the additional computational overhead result in scalability issues for MagNet and Framelet, leading to OOM errors when dealing with the billion-level dataset. 
    Although MGC decouples the graph propagation, its advantages require multiple propagations to fully manifest, leading to incomplete training within 12 hours and resulting in OOT errors.
    For detailed algorithmic complexity analysis, please refer to Appendix~\ref{appendix: MAP and MAP++ Algorithm and Complexity Analysis}.

\noindent
\textbf{A New MagDG.}
    To answer \textbf{Q2}, we present the experimental results in Table~\ref{tab: map++ overall performance} and observe that MAP++ consistently outperforms all baselines. 
    Notably, we do not conduct additional evaluations of MAP++ on link-level downstream tasks. 
    This is because, as shown in Table~\ref{tab: map link performance improvement}, performance improvements are already anticipated. 
    Given the limited space, we prioritized incorporating more SOTA undirected GNNs to ensure a fair comparison.
    However, their reliance on symmetric message-passing limits the recognition of complex directed relationships, leading to sub-optimal performance.

\begin{table}  % 表格右浮动，占据半栏宽度
\setlength{\abovecaptionskip}{0.2cm}
\setlength{\belowcaptionskip}{-0.2cm}
\caption{Node-C (ACC) improvement.
}
\label{tab: map performance improvement}
\begin{tabular}{c|cccc|c}
\midrule[0.3pt]
Models          & Actor    & Empire  & arXiv & Papers & Improv.                 \\ \midrule[0.3pt]
MagNet          & 32.4±0.5 & 78.5±0.4 & 64.5±0.6   & OOM             & \multirow{2}{*}{\textcolor{black}{$\Uparrow$4.28$\%$}} \\
+MAP            & 34.0±0.4 & 82.8±0.4 & 68.0±0.4   & OOM             &                         \\ \midrule[0.3pt]
MGC             & 33.9±0.5 & 79.1±0.3 & 63.8±0.1   & OOT             & \multirow{2}{*}{\textcolor{black}{$\Uparrow$4.96$\%$}} \\
+MAP            & 35.2±0.3 & 82.8±0.4 & 67.6±0.2   & OOT             &                         \\ \midrule[0.3pt]
Framelet        & 33.1±0.6 & 79.8±0.3 & 64.7±0.1   & OOM             & \multirow{2}{*}{\textcolor{black}{$\Uparrow$4.54$\%$}}   \\
+MAP            & 34.8±0.6 & 83.6±0.2 & 68.4±0.2   & OOM             &                         \\ \midrule[0.3pt]
LightDiC        & 33.6±0.4 & 78.8±0.2 & 65.6±0.2   & 65.4±0.2        & \multirow{2}{*}{\textcolor{black}{$\Uparrow$5.12$\%$}}   \\
+MAP            & 35.5±0.4 & 83.0±0.3 & 69.1±0.1   & 68.7±0.3        &                         \\ \midrule[0.3pt]
\end{tabular}
\vspace{-0.2cm}
\end{table}

\begin{table}  % 表格右浮动，占据半栏宽度
\setlength{\abovecaptionskip}{0.2cm}
\setlength{\belowcaptionskip}{-0.2cm}
\caption{Existence (AUC) and Direction (AP) improvement.
}
\label{tab: map link performance improvement}
\begin{tabular}{c|cc|cc|c}
\midrule[0.3pt]
Datasets & \multicolumn{2}{c|}{Slashdot (Link)} & \multicolumn{2}{c|}{Epinions(Link)} & \multirow{2}{*}{Improv.} \\
Tasks    & Exist.     & Direct.     & Exist.     & Direct.     &                          \\\midrule[0.3pt]
MagNet   & 90.3±0.1      & 92.4±0.1      & 91.6±0.0      & 91.5±0.1      & \multirow{2}{*}{$\Uparrow$2.76$\%$}    \\
+MAP     & 92.1±0.0      & 93.2±0.1      & 93.2±0.1      & 93.4±0.1      &                          \\ \midrule[0.3pt]
MGC      & 90.1±0.1      & 92.3±0.1      & 91.8±0.1      & 91.4±0.0      & \multirow{2}{*}{$\Uparrow$2.39$\%$}    \\
+MAP     & 91.9±0.1      & 93.4±0.0      & 93.0±0.0      & 93.0±0.1      &                          \\ \midrule[0.3pt]
Framelet & 90.5±0.0      & 92.5±0.1      & 91.5±0.1      & 91.0±0.1      & \multirow{2}{*}{$\Uparrow$2.46$\%$}    \\
+MAP     & 92.3±0.1      & 93.1±0.0      & 93.3±0.1      & 93.1±0.1      &                          \\ \midrule[0.3pt]
LightDiC & 90.2±0.1      & 92.4±0.0      & 91.6±0.0      & 91.2±0.1      & \multirow{2}{*}{$\Uparrow$2.81$\%$}    \\
+MAP     & 92.5±0.1      & 93.6±0.1      & 93.1±0.0      & 93.2±0.0      &                          \\ \midrule[0.3pt]
\end{tabular}
\vspace{-0.2cm}
\end{table}

\begin{table}[]
\setlength{\abovecaptionskip}{0.2cm}
\setlength{\belowcaptionskip}{-0.2cm}
\caption{Node-C (ACC) Performance.
}
\label{tab: map++ overall performance}
\begin{tabular}{ccccc}
\midrule[0.3pt]
Models         & CoraML                     & CiteSeer                      & WikiCS                        & Papers                    \\ \midrule[0.3pt]
GCNII          & 80.84±0.5                   & 62.55±0.6                      & 77.42±0.3                      & OOM          \\
GATv2          & 81.31±0.9                   & 62.82±1.0                      & 77.03±0.4                      & OOM                    \\ 
OptBG          & 81.58±0.8                   & 62.76±0.7                      & 77.58±0.5                      & 66.70±0.2                    \\
NAG            & 81.96±0.7                   & 63.12±0.8                      & 77.32±0.6                      & OOM                   \\
GAMLP          & 82.18±0.8                   & 62.94±0.9                      & 77.87±0.7                      & 66.92±0.3                    \\\midrule[0.3pt]
D-HYPR         & 81.72±0.5                   & 63.87±0.7                      & 77.76±0.2                      & OOM                     \\
HoloNet        & 81.53±06                   & 64.13±0.8                      & \underline{78.66±0.3}          & OOM                     \\ \midrule[0.3pt]
DGCN           & 81.25±0.5                   & 63.54±0.8                      & 77.44±0.3                      & OOM                     \\
DiGCN          & 81.62±0.4                   & 63.99±0.9                      & 78.41±0.6                      & OOM                       \\
NSTE           & 81.87±0.6                   & 63.63±0.7                      & 77.63±0.4                      & OOM                     \\
DIMPA          & 82.05±0.9                   & 63.14±0.9                      & 77.94±0.3                      & OOM                        \\
Dir-GNN        & 81.93±0.7                   & 64.29±0.8                      & 78.09±0.4                      & OOM                    \\
LightDiC       & 81.76±0.4                   & 64.19±0.6                      & 78.35±0.2                      & 66.83±0.2                    \\
ADPA           & \underline{82.43±0.8}       & \underline{64.50±0.9}          & 78.24±0.3                      & \underline{67.42±0.3}              \\ \midrule[0.3pt]
\textbf{MAP++} & \textbf{84.87±0.4}          & \textbf{67.58±0.8}             & \textbf{81.60±0.3}             & \textbf{69.47±0.3}         \\ \midrule[0.3pt]
\end{tabular}
\vspace{-0.2cm}
\end{table}

\subsection{Ablation Study}
\label{sec: Ablation Study}

\begin{table} % 表格右浮动，占据半栏宽度
\setlength{\abovecaptionskip}{0.2cm}
\setlength{\belowcaptionskip}{-0.2cm}
\caption{Ablation study (ACC).
}
\label{tab: ab_study}
\begin{tabular}{c|ccc}
\midrule[0.3pt]
\multirow{2}{*}{Model}                                       & CiteSeer  & Tolokers   & WikiTalk\\ 
                                     & Node-C   & Node-C   &  Link-C\\ \midrule[0.3pt]
MagNet                                      & 64.21±0.63 & 79.04±0.22 & 90.42±0.15\\
MagNet + MAP                                & 66.87±0.56 & 80.15±0.32 & 91.30±0.16\\
w/o Topology (Local)                        & 66.53±0.78 & 79.84±0.48 & 91.11±0.13\\
w/o Topology (Global)                       & 66.12±0.45 & 79.51±0.25 & 90.96±0.18\\
w/o Feature Encoding                        & 65.60±0.50 & 79.34±0.36 & 90.78±0.12\\ \midrule[0.3pt] \midrule[0.3pt]
LightDiC                                    & 63.96±0.38 & 79.18±0.19 & 90.21±0.10\\
LightDiC + MAP                              & 67.25±0.37 & 80.36±0.27 & 91.05±0.14\\
w/o Topology (Local)                        & 66.82±0.52 & 80.15±0.43 & 90.86±0.11\\
w/o Topology (Global)                       & 66.46±0.39 & 79.73±0.32 & 90.74±0.15\\
w/o Feature Encoding                        & 65.21±0.35 & 79.50±0.25 & 90.58±0.13\\ \midrule[0.3pt] \midrule[0.3pt]
MAP++                                       & 67.58±0.77 & 80.78±0.21 & 91.46±0.13\\
w/o Edge-wise Prop                               & 67.10±0.84 & 80.36±0.28 & 91.12±0.15\\
w/o Node-wise Agg                               & 66.49±0.65 & 80.12±0.24 & 90.78±0.12\\ \midrule[0.3pt]
\end{tabular}
\vspace{-0.3cm}
\end{table}

\textbf{The Key Design of MAP and MAP++.}
    To answer \textbf{Q3}, we present experimental results in Table~\ref{tab: ab_study}, evaluating the effectiveness of
    (1) Topology-related uncertainty and Feature-related correlation encoding in Sec.~\ref{sec: MAP Framework};
    (2) Edge-wise graph propagation and node-wise message aggregation in Sec.~\ref{sec: MAP++ Framework}.
    We draw the following conclusions: 
    (1) Local structural encoding models neighbor in a fine-grained manner, reducing the prediction variance of MagNet from 0.48 to 0.32 on the Tolokers.
    (2) Global structural encoding enhances performance upper bounds by regulating propagation granularity comprehensively.
    (3) Feature correlation is directly relevant to downstream tasks, thereby crucial for performance improvement. 
    Specifically, it boosts LightDiC's accuracy from 65.21 to 67.25 in CiteSeer.
    (4) Based upon these concepts, MAP++ introduces parameterized propagation kernels and attention-based message aggregation to further optimize predictions, leading to significant improvements.

\begin{figure}  % 表格右浮动，占据半栏宽度
\centering
    \setlength{\abovecaptionskip}{0.1cm}
    \setlength{\belowcaptionskip}{-0.3cm}
  \includegraphics[width=\linewidth,scale=1.00]{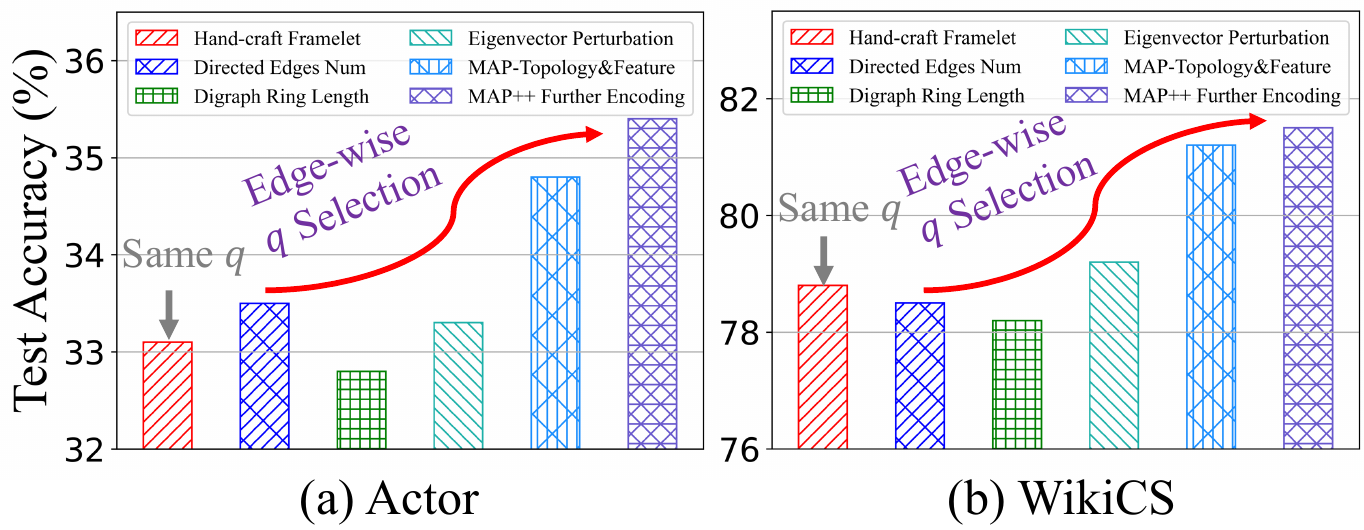}
  \caption{
    Node-C Performance with $q$ guidance.
}
  \label{fig: exp_q_selection}
  \vspace{-0.1cm}
\end{figure}

\begin{figure}[t]
	\centering
    \setlength{\abovecaptionskip}{0.1cm}
    \setlength{\belowcaptionskip}{-0.3cm}
  \includegraphics[width=\linewidth,scale=1.00]{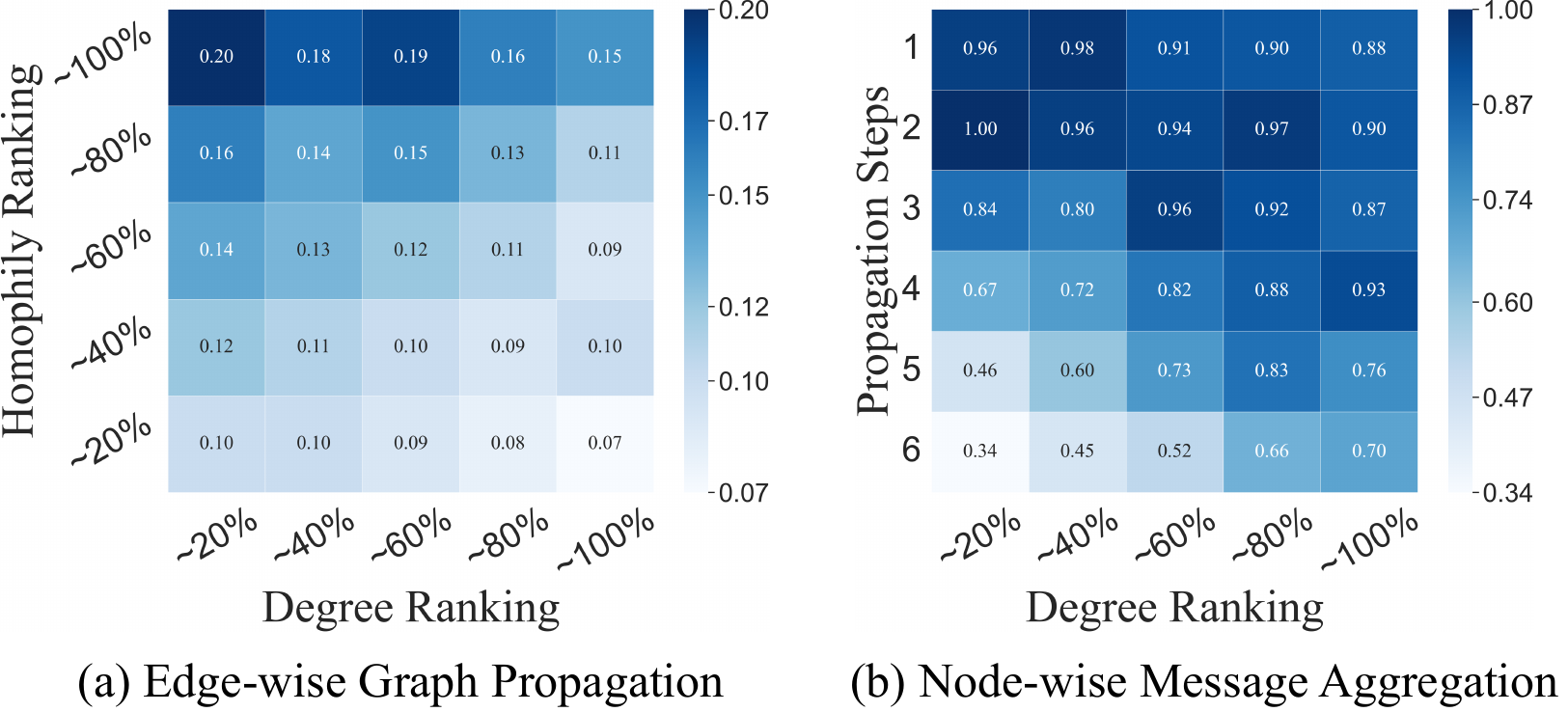}
  \caption{
    The visualization of MAP++ in arXiv.
}
    \vspace{-0.1cm}
  \label{fig: exp_vis}
\end{figure}

\begin{figure*}[t]
	\centering
     \setlength{\abovecaptionskip}{0.1cm}
    \setlength{\belowcaptionskip}{-0.3cm}
  \includegraphics[width=\textwidth]{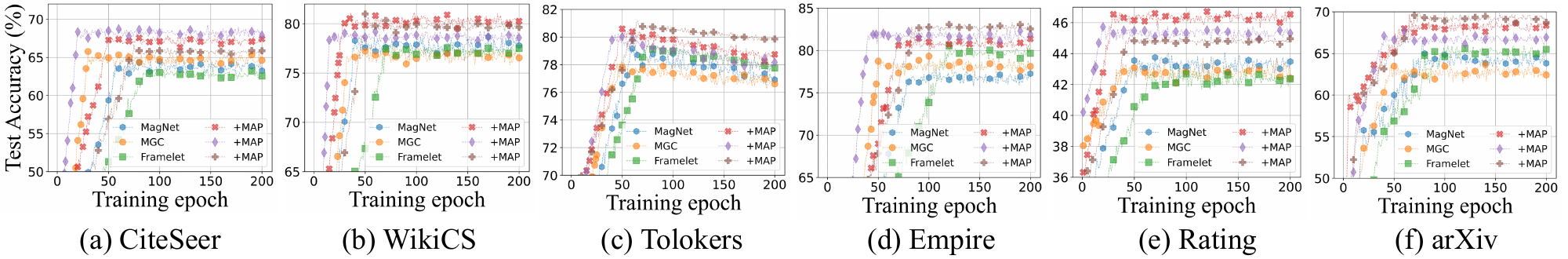}
  \caption{
    Convergence improvement brought by MAP.
}
  \label{fig: exp_converge}
\end{figure*}

\begin{figure}  % 表格右浮动，占据半栏宽度
\centering
    \setlength{\abovecaptionskip}{0.1cm}
    \setlength{\belowcaptionskip}{-0.3cm}
  \includegraphics[width=\linewidth,scale=1.00]{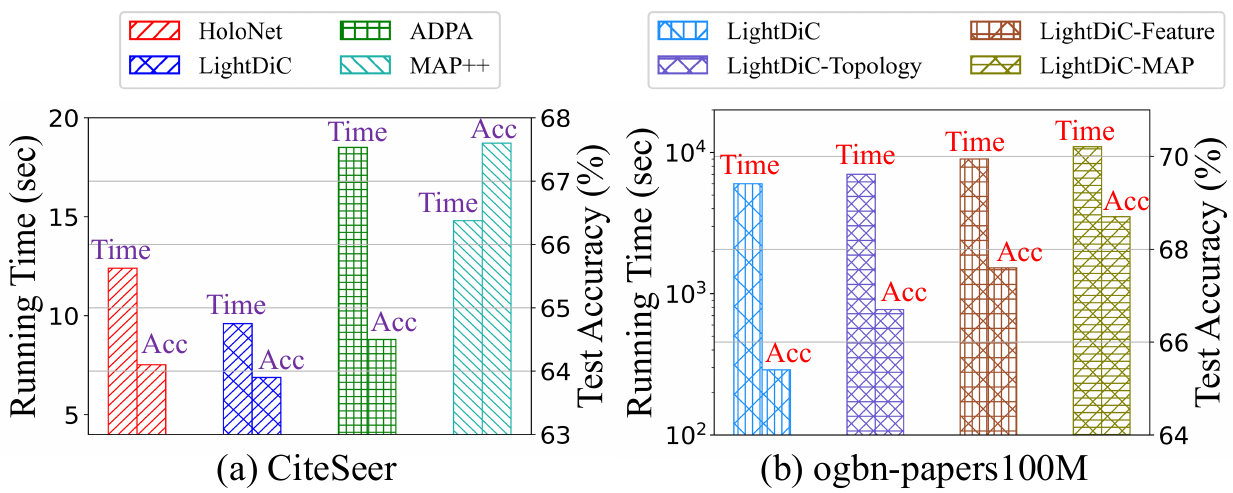}
  \caption{
    Running efficiency performance.
}
  \label{fig: exp_efficiency}
  \vspace{-0.2cm}
\end{figure}

\noindent
\textbf{$q$ Selection in Spectral Graph Theory.}
    As mentioned in Sec.~\ref{sec: Introduction}, some studies provide $q$ selection guidance from a topology perspective (see Appendix~\ref{appendix: Guidance for Selecting q in Spectral Graph Theory}).
    In this section, we review relevant studies and compare their strategies with MAP and MAP++ in the context of digraph learning shown in Fig.~\ref{fig: exp_q_selection}. 
    Drawing from experimental findings, we discern notable performance benefits exhibited by MAP and MAP++, highlighting the notion that previous approaches may not yield satisfactory results in digraph learning due to their limited incorporation of node profiles.
    Moreover, the performance of MAP++ validates the advantage of further encoding the magnetic field potentials of directed edges by learnable mechanisms.

\noindent
\textbf{The Visualization of MAP++}.
    To directly demonstrate the effectiveness of MAP++, we provide visualization in Fig.~\ref{fig: exp_vis}:
    (a) The average $q$ of directed edges within different nodes (topology-based degree ranking and feature-based homophily ranking).
    (b) The average attention weights of propagated features within different nodes (topology-based degree ranking) and propagation steps.
    Following observations validate our key insights in Sec.~\ref{sec: Empirical Investigation}:
    (1) Fig.~\ref{fig: exp_vis} (a) shows that smaller $q$ are chosen for pairs with higher-homophily, while larger $q$ are selected for pairs with higher degrees. 
    The increase in $q$ as homophily decreases underscores the importance of node attributes in digraph learning.
    (2) Fig.~\ref{fig: exp_vis} (b) shows that 1-3 step features hold significant importance, similar to 1-3 layer DiGNNs. 
    For higher-degree nodes, the weights for larger steps decrease rapidly to prevent over-smoothing by limiting irrelevant information.

\subsection{Efficiency Comparison}
\label{sec: Efficiency Comparison}
\textbf{Convergence Improvement.}
    To answer \textbf{Q4}, we present the experimental results in Fig.~\ref{fig: exp_converge}, where we observe that MAP significantly aids existing MagDGs in achieving faster and more stable convergence, along with higher accuracy.
    For instance, in WikiCS, MAP assists MGC in achieving rapid convergence around the 20th epoch, saving nearly half of the training cost.
    Notably, due to the sparse node features in Tolokers and the intricate topology in large-scale arXiv, all methods inevitably suffer from over-fitting issues and slow convergence. 
    However, integrating MAP significantly enhances the training efficiency of all baselines and mitigates these issues.

\noindent
\textbf{Runtime Overhead.}
    We provide an efficiency visualization in Fig.~\ref{fig: exp_efficiency}. 
    Despite the additional computational cost introduced by MAP for fine-grained graph propagation, the time overhead remains within acceptable limits and brings considerable performance improvement. 
    This is facilitated by topology-related one-step pre-processing and intermittent feature-related encoding during training.
    Meanwhile, while MAP++ introduces extra trainable parameters, its overall time overhead remains lower than the most competitive ADPA, thanks to its decoupled design. 
    Moreover, it exhibits significant performance advantages compared to other baselines.

\subsection{Performance under Sparse Scenarios}
\label{sec: Performance under Sparse Sraphs}

\begin{figure} % 表格右浮动，占据半栏宽度
\centering
    \setlength{\abovecaptionskip}{0.1cm}
    \setlength{\belowcaptionskip}{-0.3cm}
  \includegraphics[width=\linewidth,scale=1.00]{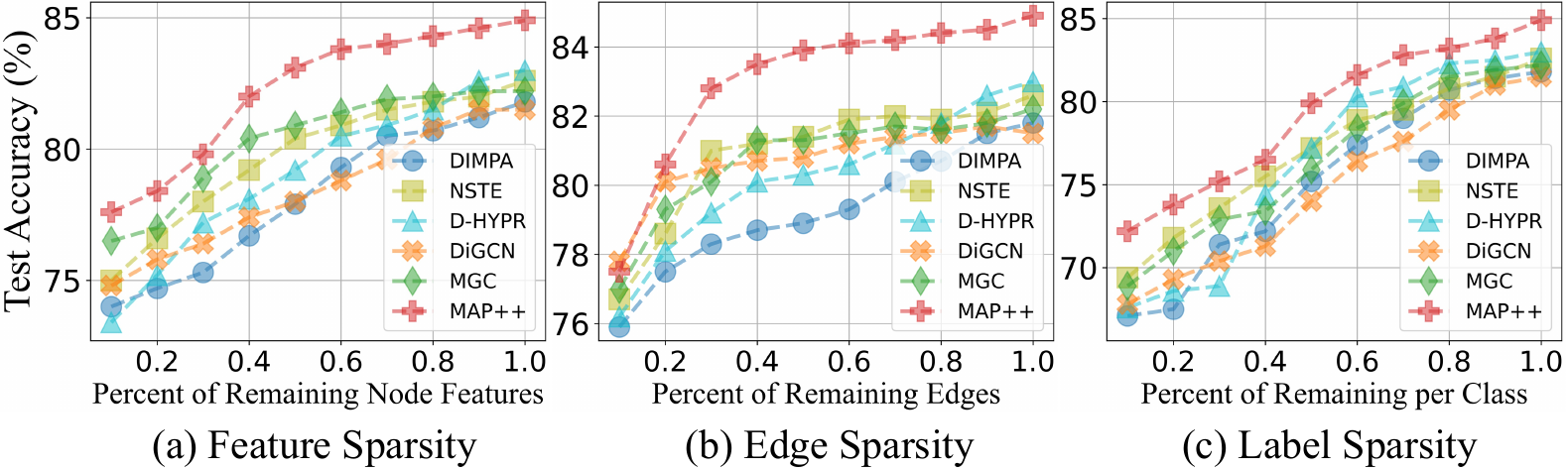}
  \caption{
    Sparsity performance on CoraML.
}
  \label{fig: exp_sparsity}
    \vspace{-0.1cm}
\end{figure}

    To answer \textbf{Q5}, we present experimental results in Fig.~\ref{fig: exp_sparsity}. 
    For feature sparsity, we introduce partial missing features for unlabeled nodes.
    Consequently, methods relying solely on node quantity, such as D-HYPR, suffer performance degradation.
    Conversely, DiGCN, MGC, and MAP++ demonstrate resilience, as their high-order propagation partially compensates for the missing features. 
    Regarding edge sparsity, since all baselines rely on topology to obtain high-quality node embeddings, they all face severe degradation. 
    However, we observe that MAP++ outperforms others due to its fine-grained message passing. 
    As for the label sparsity, we observe a similar trend to the feature sparsity. 
    These findings collectively underscore the robustness enhancements achieved by MAP++ over baselines.

\section{Conclusion}
\label{sec: Conclusion}
    In recent years, MagDGs have stood out for edge direction modeling through the complex domain, inheriting insights from undirected graph learning.
    However, the extension of the $q$-parameterized magnetic Laplacian to digraph learning remains under-explored.
    To emphasize such a research gap, we provide valuable empirical studies and theoretical analysis to obtain the $q$-parameterized criteria for digraph learning. 
    Based on this, we introduce two key techniques: MAP and MAP++. 
    The achieved SOTA performance, coupled with flexibility and scalability, serves as compelling evidence of the practicality of our approach.
    A promising direction involves tailoring complex-domain graph propagation.
    Furthermore, an in-depth analysis of magnetic potential modeling from the perspective of topological dynamics shows great potential.

\clearpage

\bibliographystyle{ACM-Reference-Format}
\bibliography{sample-base}


\begin{thebibliography}{75}

%%% the Web address.

\ifx \showCODEN    \undefined \def \showCODEN     #1{\unskip}     \fi
\ifx \showDOI      \undefined \def \showDOI       #1{#1}\fi
\ifx \showISBNx    \undefined \def \showISBNx     #1{\unskip}     \fi
\ifx \showISBNxiii \undefined \def \showISBNxiii  #1{\unskip}     \fi
\ifx \showISSN     \undefined \def \showISSN      #1{\unskip}     \fi
\ifx \showLCCN     \undefined \def \showLCCN      #1{\unskip}     \fi
\ifx \shownote     \undefined \def \shownote      #1{#1}          \fi
\ifx \showarticletitle \undefined \def \showarticletitle #1{#1}   \fi
\ifx \showURL      \undefined \def \showURL       {\relax}        \fi
% invisible to TeX
\providecommand\bibfield[2]{#2}
\providecommand\bibinfo[2]{#2}
\providecommand\natexlab[1]{#1}
\providecommand\showeprint[2][]{arXiv:#2}

\bibitem[Akiba et~al\mbox{.}(2019)]%
        {akiba2019optuna}
\bibfield{author}{\bibinfo{person}{Takuya Akiba}, \bibinfo{person}{Shotaro Sano}, \bibinfo{person}{Toshihiko Yanase}, \bibinfo{person}{Takeru Ohta}, {and} \bibinfo{person}{Masanori Koyama}.} \bibinfo{year}{2019}\natexlab{}.
\newblock \showarticletitle{Optuna: A next-generation hyperparameter optimization framework}. In \bibinfo{booktitle}{\emph{Proceedings of the ACM SIGKDD Conference on Knowledge Discovery and Data Mining, KDD}}.
\newblock


\bibitem[Bian et~al\mbox{.}(2020)]%
        {bian2020_directed_app_social1}
\bibfield{author}{\bibinfo{person}{Tian Bian}, \bibinfo{person}{Xi Xiao}, \bibinfo{person}{Tingyang Xu}, \bibinfo{person}{Peilin Zhao}, \bibinfo{person}{Wenbing Huang}, \bibinfo{person}{Yu Rong}, {and} \bibinfo{person}{Junzhou Huang}.} \bibinfo{year}{2020}\natexlab{}.
\newblock \showarticletitle{Rumor detection on social media with bi-directional graph convolutional networks}. In \bibinfo{booktitle}{\emph{Proceedings of the Association for the Advancement of Artificial Intelligence, AAAI}}.
\newblock


\bibitem[Bojchevski and Günnemann(2018)]%
        {bojchevski2018coraml_citeseer}
\bibfield{author}{\bibinfo{person}{Aleksandar Bojchevski} {and} \bibinfo{person}{Stephan Günnemann}.} \bibinfo{year}{2018}\natexlab{}.
\newblock \showarticletitle{Deep Gaussian Embedding of Graphs: Unsupervised Inductive Learning via Ranking}. In \bibinfo{booktitle}{\emph{ICLR Workshop on Representation Learning on Graphs and Manifolds}}.
\newblock


\bibitem[Brody et~al\mbox{.}(2022)]%
        {brody2021gatv2}
\bibfield{author}{\bibinfo{person}{Shaked Brody}, \bibinfo{person}{Uri Alon}, {and} \bibinfo{person}{Eran Yahav}.} \bibinfo{year}{2022}\natexlab{}.
\newblock \showarticletitle{How attentive are graph attention networks?}
\newblock \bibinfo{journal}{\emph{International Conference on Learning Representations, ICLR}} (\bibinfo{year}{2022}).
\newblock


\bibitem[Chat et~al\mbox{.}(2019)]%
        {chat2019spectral_graph_magnetic_laplacian2}
\bibfield{author}{\bibinfo{person}{Bilal~A Chat}, \bibinfo{person}{Hilal~A Ganie}, {and} \bibinfo{person}{S Pirzada}.} \bibinfo{year}{2019}\natexlab{}.
\newblock \showarticletitle{Bounds for the skew Laplacian spectral radius of oriented graphs}.
\newblock \bibinfo{journal}{\emph{Carpathian Journal of Mathematics}} \bibinfo{volume}{35}, \bibinfo{number}{1} (\bibinfo{year}{2019}), \bibinfo{pages}{31--40}.
\newblock


\bibitem[Chen et~al\mbox{.}(2023)]%
        {chen2022nagphormer}
\bibfield{author}{\bibinfo{person}{Jinsong Chen}, \bibinfo{person}{Kaiyuan Gao}, \bibinfo{person}{Gaichao Li}, {and} \bibinfo{person}{Kun He}.} \bibinfo{year}{2023}\natexlab{}.
\newblock \showarticletitle{NAGphormer: A tokenized graph transformer for node classification in large graphs}. In \bibinfo{booktitle}{\emph{International Conference on Learning Representations, ICLR}}.
\newblock


\bibitem[Chen et~al\mbox{.}(2020)]%
        {chen2020gcnii}
\bibfield{author}{\bibinfo{person}{Ming Chen}, \bibinfo{person}{Zhewei Wei}, \bibinfo{person}{Zengfeng Huang}, \bibinfo{person}{Bolin Ding}, {and} \bibinfo{person}{Yaliang Li}.} \bibinfo{year}{2020}\natexlab{}.
\newblock \showarticletitle{Simple and deep graph convolutional networks}. In \bibinfo{booktitle}{\emph{International Conference on Machine Learning, ICML}}.
\newblock


\bibitem[Chung(2005)]%
        {chung2005spectral_graph_magnetic_laplacian1}
\bibfield{author}{\bibinfo{person}{Fan Chung}.} \bibinfo{year}{2005}\natexlab{}.
\newblock \showarticletitle{Laplacians and the Cheeger inequality for directed graphs}.
\newblock \bibinfo{journal}{\emph{Annals of Combinatorics}}  \bibinfo{volume}{9} (\bibinfo{year}{2005}), \bibinfo{pages}{1--19}.
\newblock


\bibitem[Cucuringu(2016)]%
        {cucuringu2016graph_angular_synchronization3}
\bibfield{author}{\bibinfo{person}{Mihai Cucuringu}.} \bibinfo{year}{2016}\natexlab{}.
\newblock \showarticletitle{Sync-rank: Robust ranking, constrained ranking and rank aggregation via eigenvector and SDP synchronization}.
\newblock \bibinfo{journal}{\emph{IEEE Transactions on Network Science and Engineering}} \bibinfo{volume}{3}, \bibinfo{number}{1} (\bibinfo{year}{2016}), \bibinfo{pages}{58--79}.
\newblock


\bibitem[Cucuringu et~al\mbox{.}(2012)]%
        {cucuringu2012graph_angular_synchronization2}
\bibfield{author}{\bibinfo{person}{Mihai Cucuringu}, \bibinfo{person}{Yaron Lipman}, {and} \bibinfo{person}{Amit Singer}.} \bibinfo{year}{2012}\natexlab{}.
\newblock \showarticletitle{Sensor network localization by eigenvector synchronization over the Euclidean group}.
\newblock \bibinfo{journal}{\emph{ACM Transactions on Sensor Networks (TOSN)}} \bibinfo{volume}{8}, \bibinfo{number}{3} (\bibinfo{year}{2012}), \bibinfo{pages}{1--42}.
\newblock


\bibitem[Fanuel et~al\mbox{.}(2018)]%
        {fanuel2018magLaplacian1}
\bibfield{author}{\bibinfo{person}{Micha{\"e}l Fanuel}, \bibinfo{person}{Carlos~M Ala{\'\i}z}, \bibinfo{person}{{\'A}ngela Fern{\'a}ndez}, {and} \bibinfo{person}{Johan~AK Suykens}.} \bibinfo{year}{2018}\natexlab{}.
\newblock \showarticletitle{Magnetic eigenmaps for the visualization of directed networks}.
\newblock \bibinfo{journal}{\emph{Applied and Computational Harmonic Analysis}} \bibinfo{volume}{44}, \bibinfo{number}{1} (\bibinfo{year}{2018}), \bibinfo{pages}{189--199}.
\newblock


\bibitem[Fanuel et~al\mbox{.}(2017)]%
        {fanuel2017q_magnetic3}
\bibfield{author}{\bibinfo{person}{Micha{\"e}l Fanuel}, \bibinfo{person}{Carlos~M Alaiz}, {and} \bibinfo{person}{Johan~AK Suykens}.} \bibinfo{year}{2017}\natexlab{}.
\newblock \showarticletitle{Magnetic eigenmaps for community detection in directed networks}.
\newblock \bibinfo{journal}{\emph{Physical Review E}} \bibinfo{volume}{95}, \bibinfo{number}{2} (\bibinfo{year}{2017}), \bibinfo{pages}{022302}.
\newblock


\bibitem[Frasca et~al\mbox{.}(2020)]%
        {frasca2020sign}
\bibfield{author}{\bibinfo{person}{Fabrizio Frasca}, \bibinfo{person}{Emanuele Rossi}, \bibinfo{person}{Davide Eynard}, \bibinfo{person}{Ben Chamberlain}, \bibinfo{person}{Michael Bronstein}, {and} \bibinfo{person}{Federico Monti}.} \bibinfo{year}{2020}\natexlab{}.
\newblock \showarticletitle{Sign: Scalable inception graph neural networks}.
\newblock \bibinfo{journal}{\emph{arXiv preprint arXiv:2004.11198}} (\bibinfo{year}{2020}).
\newblock


\bibitem[Furutani et~al\mbox{.}(2020)]%
        {furutani2020magLaplacian2}
\bibfield{author}{\bibinfo{person}{Satoshi Furutani}, \bibinfo{person}{Toshiki Shibahara}, \bibinfo{person}{Mitsuaki Akiyama}, \bibinfo{person}{Kunio Hato}, {and} \bibinfo{person}{Masaki Aida}.} \bibinfo{year}{2020}\natexlab{}.
\newblock \showarticletitle{Graph signal processing for directed graphs based on the hermitian laplacian}. In \bibinfo{booktitle}{\emph{Joint European Conference on Machine Learning and Knowledge Discovery in Databases, ECML-PKDD}}. Springer.
\newblock


\bibitem[Geisler et~al\mbox{.}(2023)]%
        {geisler2023transformers_meet_digraph}
\bibfield{author}{\bibinfo{person}{Simon Geisler}, \bibinfo{person}{Yujia Li}, \bibinfo{person}{Daniel~J Mankowitz}, \bibinfo{person}{Ali~Taylan Cemgil}, \bibinfo{person}{Stephan G{\"u}nnemann}, {and} \bibinfo{person}{Cosmin Paduraru}.} \bibinfo{year}{2023}\natexlab{}.
\newblock \showarticletitle{Transformers meet directed graphs}. In \bibinfo{booktitle}{\emph{International Conference on Machine Learning, ICML}}. PMLR.
\newblock


\bibitem[Grave et~al\mbox{.}(2018)]%
        {grave2018fast_word_embedding}
\bibfield{author}{\bibinfo{person}{Edouard Grave}, \bibinfo{person}{Piotr Bojanowski}, \bibinfo{person}{Prakhar Gupta}, \bibinfo{person}{Armand Joulin}, {and} \bibinfo{person}{Tomas Mikolov}.} \bibinfo{year}{2018}\natexlab{}.
\newblock \showarticletitle{Learning word vectors for 157 languages}.
\newblock \bibinfo{journal}{\emph{arXiv preprint arXiv:1802.06893}} (\bibinfo{year}{2018}).
\newblock


\bibitem[Griffiths and Schroeter(2018)]%
        {griffiths2018graph_angular_synchronization6}
\bibfield{author}{\bibinfo{person}{David~J Griffiths} {and} \bibinfo{person}{Darrell~F Schroeter}.} \bibinfo{year}{2018}\natexlab{}.
\newblock \bibinfo{booktitle}{\emph{Introduction to quantum mechanics}}.
\newblock \bibinfo{publisher}{Cambridge university press}.
\newblock


\bibitem[Guo and Wei(2023)]%
        {OptBasisGNN}
\bibfield{author}{\bibinfo{person}{Yuhe Guo} {and} \bibinfo{person}{Zhewei Wei}.} \bibinfo{year}{2023}\natexlab{}.
\newblock \showarticletitle{Graph Neural Networks with Learnable and Optimal Polynomial Bases}.
\newblock  (\bibinfo{year}{2023}).
\newblock


\bibitem[Hamilton et~al\mbox{.}(2017)]%
        {hamilton2017graphsage}
\bibfield{author}{\bibinfo{person}{Will Hamilton}, \bibinfo{person}{Zhitao Ying}, {and} \bibinfo{person}{Jure Leskovec}.} \bibinfo{year}{2017}\natexlab{}.
\newblock \showarticletitle{Inductive representation learning on large graphs}.
\newblock \bibinfo{journal}{\emph{Advances in Neural Information Processing Systems, NeurIPS}} (\bibinfo{year}{2017}).
\newblock


\bibitem[He et~al\mbox{.}(2022a)]%
        {he2022msgnn}
\bibfield{author}{\bibinfo{person}{Yixuan He}, \bibinfo{person}{Michael Perlmutter}, \bibinfo{person}{Gesine Reinert}, {and} \bibinfo{person}{Mihai Cucuringu}.} \bibinfo{year}{2022}\natexlab{a}.
\newblock \showarticletitle{Msgnn: A spectral graph neural network based on a novel magnetic signed laplacian}. In \bibinfo{booktitle}{\emph{Learning on Graphs Conference, LoG}}.
\newblock


\bibitem[He et~al\mbox{.}(2022b)]%
        {he2022dimpa}
\bibfield{author}{\bibinfo{person}{Yixuan He}, \bibinfo{person}{Gesine Reinert}, {and} \bibinfo{person}{Mihai Cucuringu}.} \bibinfo{year}{2022}\natexlab{b}.
\newblock \showarticletitle{DIGRAC: Digraph Clustering Based on Flow Imbalance}. In \bibinfo{booktitle}{\emph{Learning on Graphs Conference, LoG}}.
\newblock


\bibitem[He et~al\mbox{.}(2024)]%
        {he2023graph_angular_synchronization4}
\bibfield{author}{\bibinfo{person}{Yixuan He}, \bibinfo{person}{Gesine Reinert}, \bibinfo{person}{David Wipf}, {and} \bibinfo{person}{Mihai Cucuringu}.} \bibinfo{year}{2024}\natexlab{}.
\newblock \showarticletitle{Robust angular synchronization via directed graph neural networks}.
\newblock \bibinfo{journal}{\emph{International Conference on Learning Representations, ICLR}} (\bibinfo{year}{2024}).
\newblock


\bibitem[He et~al\mbox{.}(2023)]%
        {he2023pygsd}
\bibfield{author}{\bibinfo{person}{Yixuan He}, \bibinfo{person}{Xitong Zhang}, \bibinfo{person}{Junjie Huang}, \bibinfo{person}{Benedek Rozemberczki}, \bibinfo{person}{Mihai Cucuringu}, {and} \bibinfo{person}{Gesine Reinert}.} \bibinfo{year}{2023}\natexlab{}.
\newblock \showarticletitle{PyTorch Geometric Signed Directed: A Software Package on Graph Neural Networks for Signed and Directed Graphs}. In \bibinfo{booktitle}{\emph{Learning on Graphs Conference, LoG}}. PMLR.
\newblock


\bibitem[Horn and Johnson(2012)]%
        {horn2012graph_angular_synchronization5}
\bibfield{author}{\bibinfo{person}{Roger~A Horn} {and} \bibinfo{person}{Charles~R Johnson}.} \bibinfo{year}{2012}\natexlab{}.
\newblock \bibinfo{booktitle}{\emph{Matrix analysis}}.
\newblock \bibinfo{publisher}{Cambridge university press}.
\newblock


\bibitem[Hu et~al\mbox{.}(2020)]%
        {hu2020ogb}
\bibfield{author}{\bibinfo{person}{Weihua Hu}, \bibinfo{person}{Matthias Fey}, \bibinfo{person}{Marinka Zitnik}, \bibinfo{person}{Yuxiao Dong}, \bibinfo{person}{Hongyu Ren}, \bibinfo{person}{Bowen Liu}, \bibinfo{person}{Michele Catasta}, {and} \bibinfo{person}{Jure Leskovec}.} \bibinfo{year}{2020}\natexlab{}.
\newblock \showarticletitle{Open graph benchmark: Datasets for machine learning on graphs}.
\newblock \bibinfo{journal}{\emph{Advances in Neural Information Processing Systems, NeurIPS}} (\bibinfo{year}{2020}).
\newblock


\bibitem[Huang et~al\mbox{.}(2021)]%
        {huang2020cands}
\bibfield{author}{\bibinfo{person}{Qian Huang}, \bibinfo{person}{Horace He}, \bibinfo{person}{Abhay Singh}, \bibinfo{person}{Ser-Nam Lim}, {and} \bibinfo{person}{Austin~R Benson}.} \bibinfo{year}{2021}\natexlab{}.
\newblock \showarticletitle{{Combining label propagation and simple models out-performs graph neural networks}}.
\newblock \bibinfo{journal}{\emph{International Conference on Learning Representations, ICLR}} (\bibinfo{year}{2021}).
\newblock


\bibitem[Kipf and Welling(2017)]%
        {kipf2016gcn}
\bibfield{author}{\bibinfo{person}{Thomas~N Kipf} {and} \bibinfo{person}{Max Welling}.} \bibinfo{year}{2017}\natexlab{}.
\newblock \showarticletitle{Semi-supervised classification with graph convolutional networks}. In \bibinfo{booktitle}{\emph{International Conference on Learning Representations, ICLR}}.
\newblock


\bibitem[Koke and Cremers(2023)]%
        {koke2023holonets}
\bibfield{author}{\bibinfo{person}{Christian Koke} {and} \bibinfo{person}{Daniel Cremers}.} \bibinfo{year}{2023}\natexlab{}.
\newblock \showarticletitle{HoloNets: Spectral Convolutions do extend to Directed Graphs}.
\newblock \bibinfo{journal}{\emph{arXiv preprint arXiv:2310.02232}} (\bibinfo{year}{2023}).
\newblock


\bibitem[Kollias et~al\mbox{.}(2022a)]%
        {kollias2022digae}
\bibfield{author}{\bibinfo{person}{Georgios Kollias}, \bibinfo{person}{Vasileios Kalantzis}, \bibinfo{person}{Tsuyoshi Id{\'e}}, \bibinfo{person}{Aur{\'e}lie Lozano}, {and} \bibinfo{person}{Naoki Abe}.} \bibinfo{year}{2022}\natexlab{a}.
\newblock \showarticletitle{Directed graph auto-encoders}. In \bibinfo{booktitle}{\emph{Proceedings of the Association for the Advancement of Artificial Intelligence, AAAI}}.
\newblock


\bibitem[Kollias et~al\mbox{.}(2022b)]%
        {kollias2022nste}
\bibfield{author}{\bibinfo{person}{Georgios Kollias}, \bibinfo{person}{Vasileios Kalantzis}, \bibinfo{person}{Tsuyoshi Id{\'e}}, \bibinfo{person}{Aur{\'e}lie Lozano}, {and} \bibinfo{person}{Naoki Abe}.} \bibinfo{year}{2022}\natexlab{b}.
\newblock \showarticletitle{Directed Graph Auto-Encoders}. In \bibinfo{booktitle}{\emph{Proceedings of the Association for the Advancement of Artificial Intelligence, AAAI}}.
\newblock


\bibitem[Leskovec et~al\mbox{.}(2010)]%
        {leskovec2010wikitalk}
\bibfield{author}{\bibinfo{person}{Jure Leskovec}, \bibinfo{person}{Daniel Huttenlocher}, {and} \bibinfo{person}{Jon Kleinberg}.} \bibinfo{year}{2010}\natexlab{}.
\newblock \showarticletitle{Signed networks in social media}. In \bibinfo{booktitle}{\emph{Proceedings of the SIGCHI conference on human factors in computing systems}}.
\newblock


\bibitem[Leskovec and Krevl(2014)]%
        {leskovec2014rating_original}
\bibfield{author}{\bibinfo{person}{Jure Leskovec} {and} \bibinfo{person}{Andrej Krevl}.} \bibinfo{year}{2014}\natexlab{}.
\newblock \showarticletitle{SNAP Datasets: Stanford large network dataset collection}.
\newblock  (\bibinfo{year}{2014}).
\newblock


\bibitem[Lhoest et~al\mbox{.}(2021)]%
        {lhoest2021empire_original}
\bibfield{author}{\bibinfo{person}{Quentin Lhoest}, \bibinfo{person}{Albert~Villanova del Moral}, \bibinfo{person}{Yacine Jernite}, \bibinfo{person}{Abhishek Thakur}, \bibinfo{person}{Patrick von Platen}, \bibinfo{person}{Suraj Patil}, \bibinfo{person}{Julien Chaumond}, \bibinfo{person}{Mariama Drame}, \bibinfo{person}{Julien Plu}, \bibinfo{person}{Lewis Tunstall}, {et~al\mbox{.}}} \bibinfo{year}{2021}\natexlab{}.
\newblock \showarticletitle{Datasets: A community library for natural language processing}.
\newblock \bibinfo{journal}{\emph{arXiv preprint arXiv:2109.02846}} (\bibinfo{year}{2021}).
\newblock


\bibitem[Li and Pan(2016)]%
        {li2016structural_entropy_toit_16}
\bibfield{author}{\bibinfo{person}{Angsheng Li} {and} \bibinfo{person}{Yicheng Pan}.} \bibinfo{year}{2016}\natexlab{}.
\newblock \showarticletitle{Structural information and dynamical complexity of networks}.
\newblock \bibinfo{journal}{\emph{IEEE Transactions on Information Theory}} \bibinfo{volume}{62}, \bibinfo{number}{6} (\bibinfo{year}{2016}), \bibinfo{pages}{3290--3339}.
\newblock


\bibitem[Li et~al\mbox{.}(2024a)]%
        {li2024lightdic}
\bibfield{author}{\bibinfo{person}{Xunkai Li}, \bibinfo{person}{Meihao Liao}, \bibinfo{person}{Zhengyu Wu}, \bibinfo{person}{Daohan Su}, \bibinfo{person}{Wentao Zhang}, \bibinfo{person}{Rong-Hua Li}, {and} \bibinfo{person}{Guoren Wang}.} \bibinfo{year}{2024}\natexlab{a}.
\newblock \showarticletitle{LightDiC: A Simple Yet Effective Approach for Large-Scale Digraph Representation Learning}.
\newblock \bibinfo{journal}{\emph{Proceedings of the VLDB Endowment}} (\bibinfo{year}{2024}).
\newblock


\bibitem[Li et~al\mbox{.}(2024b)]%
        {li2024_atp}
\bibfield{author}{\bibinfo{person}{Xunkai Li}, \bibinfo{person}{Jingyuan Ma}, \bibinfo{person}{Zhengyu Wu}, \bibinfo{person}{Daohan Su}, \bibinfo{person}{Wentao Zhang}, \bibinfo{person}{Rong-Hua Li}, {and} \bibinfo{person}{Guoren Wang}.} \bibinfo{year}{2024}\natexlab{b}.
\newblock \showarticletitle{Rethinking Node-wise Propagation for Large-scale Graph Learning}. In \bibinfo{booktitle}{\emph{Proceedings of the ACM Web Conference, WWW}}.
\newblock


\bibitem[Liang et~al\mbox{.}(2023)]%
        {liang2023hetdag}
\bibfield{author}{\bibinfo{person}{Jiaxuan Liang}, \bibinfo{person}{Jun Wang}, \bibinfo{person}{Guoxian Yu}, \bibinfo{person}{Wei Guo}, \bibinfo{person}{Carlotta Domeniconi}, {and} \bibinfo{person}{Maozu Guo}.} \bibinfo{year}{2023}\natexlab{}.
\newblock \showarticletitle{Directed acyclic graph learning on attributed heterogeneous network}.
\newblock \bibinfo{journal}{\emph{IEEE Transactions on Knowledge and Data Engineering}} (\bibinfo{year}{2023}).
\newblock


\bibitem[Likhobaba et~al\mbox{.}(2023)]%
        {Tolokers_original}
\bibfield{author}{\bibinfo{person}{Daniil Likhobaba}, \bibinfo{person}{Nikita Pavlichenko}, {and} \bibinfo{person}{Dmitry Ustalov}.} \bibinfo{year}{2023}\natexlab{}.
\newblock \showarticletitle{{Toloker Graph: Interaction of Crowd Annotators}}.
\newblock  (\bibinfo{year}{2023}).
\newblock
\urldef\tempurl%
\url{https://doi.org/10.5281/zenodo.7620795}
\showDOI{\tempurl}


\bibitem[Lin and Gao(2023a)]%
        {lin2023_framelet-magnet}
\bibfield{author}{\bibinfo{person}{Lequan Lin} {and} \bibinfo{person}{Junbin Gao}.} \bibinfo{year}{2023}\natexlab{a}.
\newblock \showarticletitle{A Magnetic Framelet-Based Convolutional Neural Network for Directed Graphs}. In \bibinfo{booktitle}{\emph{IEEE International Conference on Acoustics, Speech and Signal Processing, ICASSP}}.
\newblock


\bibitem[Lin and Gao(2023b)]%
        {lin2023framelet_gnn}
\bibfield{author}{\bibinfo{person}{Lequan Lin} {and} \bibinfo{person}{Junbin Gao}.} \bibinfo{year}{2023}\natexlab{b}.
\newblock \showarticletitle{A Magnetic Framelet-Based Convolutional Neural Network for Directed Graphs}. In \bibinfo{booktitle}{\emph{IEEE International Conference on Acoustics, Speech and Signal Processing, ICASSP}}.
\newblock


\bibitem[Luan et~al\mbox{.}(2022)]%
        {luan2022hete_gnn_survey2}
\bibfield{author}{\bibinfo{person}{Sitao Luan}, \bibinfo{person}{Chenqing Hua}, \bibinfo{person}{Qincheng Lu}, \bibinfo{person}{Jiaqi Zhu}, \bibinfo{person}{Mingde Zhao}, \bibinfo{person}{Shuyuan Zhang}, \bibinfo{person}{Xiao-Wen Chang}, {and} \bibinfo{person}{Doina Precup}.} \bibinfo{year}{2022}\natexlab{}.
\newblock \showarticletitle{Revisiting heterophily for graph neural networks}.
\newblock \bibinfo{journal}{\emph{Advances in Neural Information Processing Systems, NeurIPS}} (\bibinfo{year}{2022}).
\newblock


\bibitem[Luo et~al\mbox{.}(2023)]%
        {luo2024dagformer}
\bibfield{author}{\bibinfo{person}{Yuankai Luo}, \bibinfo{person}{Veronika Thost}, {and} \bibinfo{person}{Lei Shi}.} \bibinfo{year}{2023}\natexlab{}.
\newblock \showarticletitle{Transformers over Directed Acyclic Graphs}.
\newblock \bibinfo{journal}{\emph{Advances in Neural Information Processing Systems, NeurIPS}} (\bibinfo{year}{2023}).
\newblock


\bibitem[Ma et~al\mbox{.}(2021)]%
        {ma2021hete_gnn_survey1}
\bibfield{author}{\bibinfo{person}{Yao Ma}, \bibinfo{person}{Xiaorui Liu}, \bibinfo{person}{Neil Shah}, {and} \bibinfo{person}{Jiliang Tang}.} \bibinfo{year}{2021}\natexlab{}.
\newblock \showarticletitle{Is homophily a necessity for graph neural networks?}
\newblock \bibinfo{journal}{\emph{International Conference on Learning Representations, ICLR}} (\bibinfo{year}{2021}).
\newblock


\bibitem[Ma et~al\mbox{.}(2024)]%
        {ma2024dhgnn}
\bibfield{author}{\bibinfo{person}{Zitong Ma}, \bibinfo{person}{Wenbo Zhao}, {and} \bibinfo{person}{Zhe Yang}.} \bibinfo{year}{2024}\natexlab{}.
\newblock \showarticletitle{Directed Hypergraph Representation Learning for Link Prediction}. In \bibinfo{booktitle}{\emph{International Conference on Artificial Intelligence and Statistics, AISTATS}}.
\newblock


\bibitem[Maekawa et~al\mbox{.}(2023)]%
        {maekawa2023a2dug}
\bibfield{author}{\bibinfo{person}{Seiji Maekawa}, \bibinfo{person}{Yuya Sasaki}, {and} \bibinfo{person}{Makoto Onizuka}.} \bibinfo{year}{2023}\natexlab{}.
\newblock \showarticletitle{Why Using Either Aggregated Features or Adjacency Lists in Directed or Undirected Graph? Empirical Study and Simple Classification Method}.
\newblock \bibinfo{journal}{\emph{arXiv preprint arXiv:2306.08274}} (\bibinfo{year}{2023}).
\newblock


\bibitem[Massa and Avesani(2005)]%
        {massa2005epinions}
\bibfield{author}{\bibinfo{person}{Paolo Massa} {and} \bibinfo{person}{Paolo Avesani}.} \bibinfo{year}{2005}\natexlab{}.
\newblock \showarticletitle{Controversial users demand local trust metrics: An experimental study on epinions. com community}. In \bibinfo{booktitle}{\emph{Proceedings of the Association for the Advancement of Artificial Intelligence, AAAI}}.
\newblock


\bibitem[Mernyei and Cangea(2020)]%
        {mernyei2020wikics}
\bibfield{author}{\bibinfo{person}{P{\'e}ter Mernyei} {and} \bibinfo{person}{C{\u{a}}t{\u{a}}lina Cangea}.} \bibinfo{year}{2020}\natexlab{}.
\newblock \showarticletitle{Wiki-CS: A Wikipedia-Based Benchmark for Graph Neural Networks}.
\newblock \bibinfo{journal}{\emph{arXiv preprint arXiv:2007.02901}} (\bibinfo{year}{2020}).
\newblock


\bibitem[Ngo(2005)]%
        {ngo2005eigenvalue_perturbation_theory}
\bibfield{author}{\bibinfo{person}{Khiem~V Ngo}.} \bibinfo{year}{2005}\natexlab{}.
\newblock \showarticletitle{An approach of eigenvalue perturbation theory}.
\newblock \bibinfo{journal}{\emph{Applied Numerical Analysis \& Computational Mathematics}} \bibinfo{volume}{2}, \bibinfo{number}{1} (\bibinfo{year}{2005}), \bibinfo{pages}{108--125}.
\newblock


\bibitem[Ordozgoiti et~al\mbox{.}(2020)]%
        {ordozgoiti2020slashdot}
\bibfield{author}{\bibinfo{person}{Bruno Ordozgoiti}, \bibinfo{person}{Antonis Matakos}, {and} \bibinfo{person}{Aristides Gionis}.} \bibinfo{year}{2020}\natexlab{}.
\newblock \showarticletitle{Finding large balanced subgraphs in signed networks}. In \bibinfo{booktitle}{\emph{Proceedings of the ACM Web Conference, WWW}}.
\newblock


\bibitem[Pearson(1905)]%
        {pearson1905random_walk}
\bibfield{author}{\bibinfo{person}{Karl Pearson}.} \bibinfo{year}{1905}\natexlab{}.
\newblock \showarticletitle{The problem of the random walk}.
\newblock \bibinfo{journal}{\emph{Nature}} \bibinfo{volume}{72}, \bibinfo{number}{1865} (\bibinfo{year}{1905}), \bibinfo{pages}{294--294}.
\newblock


\bibitem[Pei et~al\mbox{.}(2020)]%
        {pei2020geomgcn}
\bibfield{author}{\bibinfo{person}{Hongbin Pei}, \bibinfo{person}{Bingzhe Wei}, \bibinfo{person}{Kevin Chen-Chuan Chang}, \bibinfo{person}{Yu Lei}, {and} \bibinfo{person}{Bo Yang}.} \bibinfo{year}{2020}\natexlab{}.
\newblock \showarticletitle{Geom-gcn: Geometric graph convolutional networks}. In \bibinfo{booktitle}{\emph{International Conference on Learning Representations, ICLR}}.
\newblock


\bibitem[Pennington et~al\mbox{.}(2014)]%
        {pennington2014glove}
\bibfield{author}{\bibinfo{person}{Jeffrey Pennington}, \bibinfo{person}{Richard Socher}, {and} \bibinfo{person}{Christopher~D Manning}.} \bibinfo{year}{2014}\natexlab{}.
\newblock \showarticletitle{Glove: Global vectors for word representation}. In \bibinfo{booktitle}{\emph{Proceedings of Conference on Empirical Methods in Natural Language Processing, EMNLP}}.
\newblock


\bibitem[Platonov et~al\mbox{.}(2023)]%
        {platonov2023hete_gnn_survey4}
\bibfield{author}{\bibinfo{person}{Oleg Platonov}, \bibinfo{person}{Denis Kuznedelev}, \bibinfo{person}{Michael Diskin}, \bibinfo{person}{Artem Babenko}, {and} \bibinfo{person}{Liudmila Prokhorenkova}.} \bibinfo{year}{2023}\natexlab{}.
\newblock \showarticletitle{A critical look at the evaluation of GNNs under heterophily: are we really making progress?}
\newblock \bibinfo{journal}{\emph{International Conference on Learning Representations, ICLR}} (\bibinfo{year}{2023}).
\newblock


\bibitem[Rossi et~al\mbox{.}(2023)]%
        {dirgnn_rossi_2023}
\bibfield{author}{\bibinfo{person}{Emanuele Rossi}, \bibinfo{person}{Bertrand Charpentier}, \bibinfo{person}{Francesco~Di Giovanni}, \bibinfo{person}{Fabrizio Frasca}, \bibinfo{person}{Stephan Günnemann}, {and} \bibinfo{person}{Michael Bronstein}.} \bibinfo{year}{2023}\natexlab{}.
\newblock \showarticletitle{Edge Directionality Improves Learning on Heterophilic Graphs}.
\newblock \bibinfo{journal}{\emph{in Proceedings of The European Conference on Machine Learning and Principles and Practice of Knowledge Discovery in Databases, ECML-PKDD Workshop}} (\bibinfo{year}{2023}).
\newblock


\bibitem[Schweimer et~al\mbox{.}(2022)]%
        {schweimer2022_directed_app_social2}
\bibfield{author}{\bibinfo{person}{Christoph Schweimer}, \bibinfo{person}{Christine Gfrerer}, \bibinfo{person}{Florian Lugstein}, \bibinfo{person}{David Pape}, \bibinfo{person}{Jan~A Velimsky}, \bibinfo{person}{Robert Els{\"a}sser}, {and} \bibinfo{person}{Bernhard~C Geiger}.} \bibinfo{year}{2022}\natexlab{}.
\newblock \showarticletitle{Generating simple directed social network graphs for information spreading}. In \bibinfo{booktitle}{\emph{Proceedings of the ACM Web Conference, WWW}}.
\newblock


\bibitem[Singer(2011)]%
        {singer2011graph_angular_synchronization1}
\bibfield{author}{\bibinfo{person}{Amit Singer}.} \bibinfo{year}{2011}\natexlab{}.
\newblock \showarticletitle{Angular synchronization by eigenvectors and semidefinite programming}.
\newblock \bibinfo{journal}{\emph{Applied and computational harmonic analysis}} \bibinfo{volume}{30}, \bibinfo{number}{1} (\bibinfo{year}{2011}), \bibinfo{pages}{20--36}.
\newblock


\bibitem[Song et~al\mbox{.}(2022)]%
        {song2022gnn_survey4}
\bibfield{author}{\bibinfo{person}{Zixing Song}, \bibinfo{person}{Xiangli Yang}, \bibinfo{person}{Zenglin Xu}, {and} \bibinfo{person}{Irwin King}.} \bibinfo{year}{2022}\natexlab{}.
\newblock \showarticletitle{Graph-based semi-supervised learning: A comprehensive review}.
\newblock \bibinfo{journal}{\emph{IEEE Transactions on Neural Networks and Learning Systems}} (\bibinfo{year}{2022}).
\newblock


\bibitem[Sun et~al\mbox{.}(2021)]%
        {sun2021sagn}
\bibfield{author}{\bibinfo{person}{Chuxiong Sun}, \bibinfo{person}{Hongming Gu}, {and} \bibinfo{person}{Jie Hu}.} \bibinfo{year}{2021}\natexlab{}.
\newblock \showarticletitle{Scalable and adaptive graph neural networks with self-label-enhanced training}.
\newblock \bibinfo{journal}{\emph{arXiv preprint arXiv:2104.09376}} (\bibinfo{year}{2021}).
\newblock


\bibitem[Sun et~al\mbox{.}(2024)]%
        {sun2023adpa}
\bibfield{author}{\bibinfo{person}{Henan Sun}, \bibinfo{person}{Xunkai Li}, \bibinfo{person}{Zhengyu Wu}, \bibinfo{person}{Daohan Su}, \bibinfo{person}{Rong-Hua Li}, {and} \bibinfo{person}{Guoren Wang}.} \bibinfo{year}{2024}\natexlab{}.
\newblock \showarticletitle{Breaking the Entanglement of Homophily and Heterophily in Semi-supervised Node Classification}. In \bibinfo{booktitle}{\emph{International Conference on Data Engineering, ICDE}}.
\newblock


\bibitem[Thost and Chen(2021)]%
        {thost2021dagnn}
\bibfield{author}{\bibinfo{person}{Veronika Thost} {and} \bibinfo{person}{Jie Chen}.} \bibinfo{year}{2021}\natexlab{}.
\newblock \showarticletitle{Directed acyclic graph neural networks}.
\newblock \bibinfo{journal}{\emph{arXiv preprint arXiv:2101.07965}} (\bibinfo{year}{2021}).
\newblock


\bibitem[Tong et~al\mbox{.}(2020a)]%
        {tong2020digcn}
\bibfield{author}{\bibinfo{person}{Zekun Tong}, \bibinfo{person}{Yuxuan Liang}, \bibinfo{person}{Changsheng Sun}, \bibinfo{person}{Xinke Li}, \bibinfo{person}{David Rosenblum}, {and} \bibinfo{person}{Andrew Lim}.} \bibinfo{year}{2020}\natexlab{a}.
\newblock \showarticletitle{Digraph inception convolutional networks}.
\newblock \bibinfo{journal}{\emph{Advances in Neural Information Processing Systems, NeurIPS}} (\bibinfo{year}{2020}).
\newblock


\bibitem[Tong et~al\mbox{.}(2020b)]%
        {tong2020dgcn}
\bibfield{author}{\bibinfo{person}{Zekun Tong}, \bibinfo{person}{Yuxuan Liang}, \bibinfo{person}{Changsheng Sun}, \bibinfo{person}{David~S Rosenblum}, {and} \bibinfo{person}{Andrew Lim}.} \bibinfo{year}{2020}\natexlab{b}.
\newblock \showarticletitle{Directed graph convolutional network}.
\newblock \bibinfo{journal}{\emph{arXiv preprint arXiv:2004.13970}} (\bibinfo{year}{2020}).
\newblock


\bibitem[Veli{\v{c}}kovi{\'c} et~al\mbox{.}(2018)]%
        {velivckovic2017gat}
\bibfield{author}{\bibinfo{person}{Petar Veli{\v{c}}kovi{\'c}}, \bibinfo{person}{Guillem Cucurull}, \bibinfo{person}{Arantxa Casanova}, \bibinfo{person}{Adriana Romero}, \bibinfo{person}{Pietro Lio}, {and} \bibinfo{person}{Yoshua Bengio}.} \bibinfo{year}{2018}\natexlab{}.
\newblock \showarticletitle{Graph attention networks}. In \bibinfo{booktitle}{\emph{International Conference on Learning Representations, ICLR}}.
\newblock


\bibitem[Virinchi and Saladi(2023)]%
        {Virinchi2023_directed_app_recommendation4}
\bibfield{author}{\bibinfo{person}{Srinivas Virinchi} {and} \bibinfo{person}{Anoop Saladi}.} \bibinfo{year}{2023}\natexlab{}.
\newblock \showarticletitle{BLADE: Biased Neighborhood Sampling based Graph Neural Network for Directed Graphs}. In \bibinfo{booktitle}{\emph{Proceedings of the ACM International Conference on Web Search and Data Mining, WSDM}}.
\newblock


\bibitem[Wang et~al\mbox{.}(2020)]%
        {wang2020microsoft_MAG}
\bibfield{author}{\bibinfo{person}{Kuansan Wang}, \bibinfo{person}{Zhihong Shen}, \bibinfo{person}{Chiyuan Huang}, \bibinfo{person}{Chieh-Han Wu}, \bibinfo{person}{Yuxiao Dong}, {and} \bibinfo{person}{Anshul Kanakia}.} \bibinfo{year}{2020}\natexlab{}.
\newblock \showarticletitle{Microsoft academic graph: When experts are not enough}.
\newblock \bibinfo{journal}{\emph{Quantitative Science Studies}} \bibinfo{volume}{1}, \bibinfo{number}{1} (\bibinfo{year}{2020}), \bibinfo{pages}{396--413}.
\newblock


\bibitem[Xu et~al\mbox{.}(2018)]%
        {xu2018jknet}
\bibfield{author}{\bibinfo{person}{Keyulu Xu}, \bibinfo{person}{Chengtao Li}, \bibinfo{person}{Yonglong Tian}, \bibinfo{person}{Tomohiro Sonobe}, \bibinfo{person}{Ken-ichi Kawarabayashi}, {and} \bibinfo{person}{Stefanie Jegelka}.} \bibinfo{year}{2018}\natexlab{}.
\newblock \showarticletitle{Representation learning on graphs with jumping knowledge networks}. In \bibinfo{booktitle}{\emph{International Conference on Machine Learning, ICML}}.
\newblock


\bibitem[Zhang et~al\mbox{.}(2021b)]%
        {zhang2021mgc}
\bibfield{author}{\bibinfo{person}{Jie Zhang}, \bibinfo{person}{Bo Hui}, \bibinfo{person}{Po-Wei Harn}, \bibinfo{person}{Min-Te Sun}, {and} \bibinfo{person}{Wei-Shinn Ku}.} \bibinfo{year}{2021}\natexlab{b}.
\newblock \showarticletitle{MGC: A complex-valued graph convolutional network for directed graphs}.
\newblock \bibinfo{journal}{\emph{arXiv e-prints}} (\bibinfo{year}{2021}), \bibinfo{pages}{arXiv--2110}.
\newblock


\bibitem[Zhang et~al\mbox{.}(2021c)]%
        {zhang2021rod}
\bibfield{author}{\bibinfo{person}{Wentao Zhang}, \bibinfo{person}{Yuezihan Jiang}, \bibinfo{person}{Yang Li}, \bibinfo{person}{Zeang Sheng}, \bibinfo{person}{Yu Shen}, \bibinfo{person}{Xupeng Miao}, \bibinfo{person}{Liang Wang}, \bibinfo{person}{Zhi Yang}, {and} \bibinfo{person}{Bin Cui}.} \bibinfo{year}{2021}\natexlab{c}.
\newblock \showarticletitle{ROD: reception-aware online distillation for sparse graphs}. In \bibinfo{booktitle}{\emph{Proceedings of the ACM SIGKDD Conference on Knowledge Discovery and Data Mining, KDD}}.
\newblock


\bibitem[Zhang et~al\mbox{.}(2022)]%
        {gamlp}
\bibfield{author}{\bibinfo{person}{Wentao Zhang}, \bibinfo{person}{Ziqi Yin}, \bibinfo{person}{Zeang Sheng}, \bibinfo{person}{Yang Li}, \bibinfo{person}{Wen Ouyang}, \bibinfo{person}{Xiaosen Li}, \bibinfo{person}{Yangyu Tao}, \bibinfo{person}{Zhi Yang}, {and} \bibinfo{person}{Bin Cui}.} \bibinfo{year}{2022}\natexlab{}.
\newblock \showarticletitle{Graph Attention Multi-Layer Perceptron}.
\newblock \bibinfo{journal}{\emph{Proceedings of the ACM SIGKDD Conference on Knowledge Discovery and Data Mining, KDD}} (\bibinfo{year}{2022}).
\newblock


\bibitem[Zhang et~al\mbox{.}(2021a)]%
        {zhang2021magnet}
\bibfield{author}{\bibinfo{person}{Xitong Zhang}, \bibinfo{person}{Yixuan He}, \bibinfo{person}{Nathan Brugnone}, \bibinfo{person}{Michael Perlmutter}, {and} \bibinfo{person}{Matthew Hirn}.} \bibinfo{year}{2021}\natexlab{a}.
\newblock \showarticletitle{Magnet: A neural network for directed graphs}.
\newblock \bibinfo{journal}{\emph{Advances in Neural Information Processing Systems, NeurIPS}} (\bibinfo{year}{2021}).
\newblock


\bibitem[Zhang et~al\mbox{.}(2024)]%
        {zhang2024dglp}
\bibfield{author}{\bibinfo{person}{Yusen Zhang}, \bibinfo{person}{Yusong Tan}, \bibinfo{person}{Songlei Jian}, \bibinfo{person}{Qingbo Wu}, {and} \bibinfo{person}{Kenli Li}.} \bibinfo{year}{2024}\natexlab{}.
\newblock \showarticletitle{DGLP: Incorporating Orientation Information for Enhanced Link Prediction in Directed Graphs}. In \bibinfo{booktitle}{\emph{IEEE International Conference on Acoustics, Speech and Signal Processing, ICASSP}}.
\newblock


\bibitem[Zhao et~al\mbox{.}(2021)]%
        {zhao2021ugrec_directed_app_recommendation1}
\bibfield{author}{\bibinfo{person}{Xinxiao Zhao}, \bibinfo{person}{Zhiyong Cheng}, \bibinfo{person}{Lei Zhu}, \bibinfo{person}{Jiecai Zheng}, {and} \bibinfo{person}{Xueqing Li}.} \bibinfo{year}{2021}\natexlab{}.
\newblock \showarticletitle{UGRec: modeling directed and undirected relations for recommendation}. In \bibinfo{booktitle}{\emph{Proceedings of the International ACM SIGIR Conference on Research and Development in Information Retrieval, SIGIR}}.
\newblock


\bibitem[Zheng et~al\mbox{.}(2022)]%
        {zheng2022hete_gnn_survey3}
\bibfield{author}{\bibinfo{person}{Xin Zheng}, \bibinfo{person}{Yixin Liu}, \bibinfo{person}{Shirui Pan}, \bibinfo{person}{Miao Zhang}, \bibinfo{person}{Di Jin}, {and} \bibinfo{person}{Philip~S Yu}.} \bibinfo{year}{2022}\natexlab{}.
\newblock \showarticletitle{Graph neural networks for graphs with heterophily: A survey}.
\newblock \bibinfo{journal}{\emph{arXiv preprint arXiv:2202.07082}} (\bibinfo{year}{2022}).
\newblock


\bibitem[Zhou et~al\mbox{.}(2022)]%
        {zhou2022dhypr}
\bibfield{author}{\bibinfo{person}{Honglu Zhou}, \bibinfo{person}{Advith Chegu}, \bibinfo{person}{Samuel~S Sohn}, \bibinfo{person}{Zuohui Fu}, \bibinfo{person}{Gerard De~Melo}, {and} \bibinfo{person}{Mubbasir Kapadia}.} \bibinfo{year}{2022}\natexlab{}.
\newblock \showarticletitle{D-HYPR: Harnessing Neighborhood Modeling and Asymmetry Preservation for Digraph Representation Learning}.
\newblock \bibinfo{journal}{\emph{Proceedings of the ACM International Conference on Information and Knowledge Management, CIKM}} (\bibinfo{year}{2022}).
\newblock


\bibitem[Zou et~al\mbox{.}(2024)]%
        {zou2024_svd-magnet}
\bibfield{author}{\bibinfo{person}{Chunya Zou}, \bibinfo{person}{Andi Han}, \bibinfo{person}{Lequan Lin}, \bibinfo{person}{Ming Li}, {and} \bibinfo{person}{Junbin Gao}.} \bibinfo{year}{2024}\natexlab{}.
\newblock \showarticletitle{A Simple Yet Effective Framelet-Based Graph Neural Network for Directed Graphs}.
\newblock \bibinfo{journal}{\emph{IEEE Transactions on Artificial Intelligence}} \bibinfo{volume}{5}, \bibinfo{number}{4} (\bibinfo{year}{2024}), \bibinfo{pages}{1647--1657}.
\newblock


\end{thebibliography}

\clearpage

\appendix

\section{Outline}

The appendix is organized as follows:
\begin{description}
    \item[A.1] Algorithm and Complexity Analysis.
    \item[A.2] The Proof of Theorem~\ref{theorem: 1}.
    \item[A.3] The Proof of Theorem~\ref{theorem: 2}.
    \item[A.4] The Proof of Theorem~\ref{theorem: 3}.
    \item[A.5] Our Approach and GNNSync.
    \item[A.6] Dataset Description.
    \item[A.7] Compared Baselines.
    \item[A.8] Hyperparameter Settings.
    \item[A.9] Experiment Environment.
    \item[A.10] $q$ Selection in Spectral Graph Theory.
\end{description}

\subsection{Algorithm and Complexity Analysis}
\label{appendix: MAP and MAP++ Algorithm and Complexity Analysis}

    For a more comprehensive presentation, we provide the complete algorithm of MAP and MAP++ in Algorithm~\ref{alg: map} and Algorithm~\ref{alg: map++}.

\begin{algorithm}[htbp]
    \renewcommand{\algorithmicrequire}{\textbf{Input:}}
	\renewcommand{\algorithmicensure}{\textbf{Output:}}
	\caption{Magnetic Adaptive Propagation (MAP)}
    \label{alg: map}
    \begin{algorithmic}[1]
        \STATE \uuline{Topology-related One-step Pre-process}
        \STATE Calculate node in-degrees $\mathbf{d}_{in}(i)=\sum_{j}\mathbf{A}^\top_{ij}$;
        \STATE Calculate node out-degrees $\mathbf{d}_{out}(i)=\sum_{j}\mathbf{A}_{ij}$;
        \STATE Calculate node degrees $\mathbf{d}(i)=\mathbf{d}_{out}(i)+\mathbf{d}_{in}(i)$;
        \STATE Calculate node triple motifs $\mathbf{m}(i)=\sum_{j}\left(\mathbf{A}^2\odot\mathbf{A}^\top\right)_{ij}$;
        \STATE Calculate node cluster coefficient $\mathbf{cc}(i)=\frac{\mathbf{m}(i)}{\mathbf{d}_{in}(i)\cdot\mathbf{d}_{out}(i)}$;
        \STATE Calculate the symmetric normalization Laplacian ${\mathbf{L}}_m=\tilde{\mathbf{D}}_m^{-1/2}\tilde{\mathbf{A}}_m\tilde{\mathbf{D}}_m^{-1/2}$ according to self-loop adjacency matrix $\tilde{\mathbf{A}}_m=\mathbf{A}+\mathbf{A}^\top+\mathbf{I}$ and corresponding degree matrix $\tilde{\mathbf{D}}_m(i,i)=\sum_{j}\tilde{\mathbf{A}}_m(i,j)$;
        \STATE Calculate initialized magnetic field potential encoding $\boldsymbol{\Theta}^{(1/4)}=1/2\pi\left(\mathbf{A}-\mathbf{A}^\top\right)$;
        \STATE $\mathbf{Q}^{topo}={\operatorname{Norm}}\left(\operatorname{Global-Centrality}\left(\mathbf{d}\right)+\operatorname{Local-Centrality}\left(\mathbf{cc}\right)\right)$, $\operatorname{Norm}\left(\mathbf{x}\right)={\rm tanh}\left(\frac{\mathbf{x}}{{\rm mean}(\mathbf{x})}\right)$;
        
        \STATE \uuline{Feature-related Correlation Encoding}
        \FORALL {epoch=$1,2,\cdots,E$}
                \IF {epoch \% $e \neq 0$}
                    \STATE $\star\;\text{MGO}:=\hat{\mathbf{A}}_m^{(q)}={\mathbf{L}}_m\odot\exp\left(i\boldsymbol{\Theta}^{(1/4)}\odot\mathbf{Q}^{topo}\right)$;
                \ELSE
                    \STATE $\star\;\text{MGO}:=\hat{\mathbf{A}}_m^{(q)}={\mathbf{L}}_m\odot\exp\left(i\boldsymbol{\Theta}^{(1/4)}\odot\mathbf{Q}^{topo}\odot\mathbf{Q}^{feat}\right)$;
                \ENDIF
            \FORALL {$l=1,2,\cdots,L$}
            \STATE $\mathbf{H}^{(l)}=\operatorname{MagDG}\left(\star\;\text{MGO}, \mathbf{H}^{(l-1)}, \mathbf{W}^{(l)}\right)$;
            \ENDFOR
            \STATE Calculate node soft label ${\mathbf{Z}}={\rm Softmax}\left(\mathbf{H}^{(L)}\right)$;
            \STATE Update trainable weights in the message aggregation layers $\{\mathbf{W}^{(1)},\mathbf{W}^{(2)}\cdots,\mathbf{W}^{(L)}\}$;
            \STATE Replace the soft label of the training set node with the real label in the training sets $\mathbf{Y}_{\mathcal{V}_l}$;
            \STATE $\mathbf{Q}^{feat}=\operatorname{Norm}\left(\arccos\left(\frac{\mathbf{Z}_u\cdot\mathbf{Z}_v}{\Vert\mathbf{Z}_u \Vert\times \Vert \mathbf{Z}_v\Vert}\right)\right)$, $ \operatorname{Norm}(\mathbf{x}) = \frac{2\mathbf{x}}{\pi}$;
            \STATE Calculate node predictions $\hat{\mathbf{Y}}$ by the soft label ${\mathbf{Z}}$;
        \ENDFOR
    \end{algorithmic}
\end{algorithm}

\begin{algorithm}[htbp]
    \renewcommand{\algorithmicrequire}{\textbf{Input:}}
	\renewcommand{\algorithmicensure}{\textbf{Output:}}
	\caption{Learnable Magnetic Adaptive Propagation (MAP++)}
    \label{alg: map++}
    \begin{algorithmic}[1] 
        \REQUIRE  adjacency matrix $\mathbf{A}$, feature matrix $\mathbf{X}$, training epoch $E$, intermittent feature-related correlation encoding epoch $e$, graph propagation steps $L$, message update layer $\mathbf{W}_{update}$, training set labels $\mathbf{Y}_{\mathcal{V}_l}$, and learnable $\mathbf{W}_{edge}$ and $\mathbf{W}_{node}$;
        
	\ENSURE Node predictions or link predictions $\hat{\mathbf{Y}}$ based on the node embedding obtained by edge-wise graph propagation and node-wise message aggregation;
 
        \STATE \uuline{Topology-related One-step Pre-process}
        \STATE Calculate node in-degrees $\mathbf{d}_{in}(i)=\sum_{j}\mathbf{A}^\top_{ij}$;
        \STATE Calculate node out-degrees $\mathbf{d}_{out}(i)=\sum_{j}\mathbf{A}_{ij}$;
        \STATE Calculate node degrees $\mathbf{d}(i)=\mathbf{d}_{out}(i)+\mathbf{d}_{in}(i)$;
        \STATE Calculate node triple motifs $\mathbf{m}(i)=\sum_{j}\left(\mathbf{A}^2\odot\mathbf{A}^\top\right)_{ij}$;
        \STATE Calculate node cluster coefficient $\mathbf{cc}(i)=\frac{\mathbf{m}(i)}{\mathbf{d}_{in}(i)\cdot\mathbf{d}_{out}(i)}$;
        \STATE Calculate the symmetric normalization Laplacian ${\mathbf{L}}_m=\tilde{\mathbf{D}}_m^{-1/2}\tilde{\mathbf{A}}_m\tilde{\mathbf{D}}_m^{-1/2}$ according to the self-loop adjacency matrix $\tilde{\mathbf{A}}_m=\mathbf{A}+\mathbf{A}^\top+\mathbf{I}$ and the corresponding degree matrix $\tilde{\mathbf{D}}_m(i,i)=\sum_{j}\tilde{\mathbf{A}}_m(i,j)$;
        \STATE Calculate initialized magnetic field potential encoding $\boldsymbol{\Theta}^{(1/4)}=1/2\pi\left(\mathbf{A}-\mathbf{A}^\top\right)$;
        \STATE $\mathbf{Q}^{topo}={\operatorname{Norm}}\left(\operatorname{Global-Centrality}\left(\mathbf{d}\right)+\operatorname{Local-Centrality}\left(\mathbf{cc}\right)\right)$, $\operatorname{Norm}\left(\mathbf{x}\right)={\rm tanh}\left(\frac{\mathbf{x}}{{\rm mean}(\mathbf{x})}\right)$;

        \FORALL {epoch=$1,2,\cdots,E$}
        \STATE \uuline{Edge-wise Graph Propagation (or weight-free MAP)}
        \STATE (Depending on the computational capacity and data size.)
                \IF {epoch \% $e \neq 0$}
                    \STATE Calculate $\star\;\text{MGO}:=\hat{\mathbf{A}}_m^{(q)}=\hat{\mathbf{L}}_m\odot\exp\left(i\boldsymbol{\Theta}^{(1/4)}\odot\mathbf{Q}^{topo}\right)$;
                \ELSE
                    \STATE $\operatorname{temp}=$
                    \STATE $\operatorname{Edge-Mag}\left({\operatorname{Norm}}\left(GC\left(\mathbf{d}\right)\odot\mathbf{Q}^{feat}\Vert LC\left(\mathbf{cc}\right)\odot\mathbf{Q}^{feat}\right)\right)$, $\operatorname{Norm}\left(\mathbf{x}\right)={\rm tanh}\left(\frac{\mathbf{x}}{{\rm mean}(\mathbf{x})}\right)$;
                    \STATE Calculate $\star\;\text{MGO}:=\hat{\mathbf{A}}_m^{(q)}={\mathbf{L}}_m\odot\exp\left(i\boldsymbol{\Theta}^{(1/4)}\odot \operatorname{temp}\right)$;
                \ENDIF
            \STATE $\widetilde{\mathbf{X}}^{(L)}=\hat{\mathbf{A}}_m^{\star L}\widetilde{\mathbf{X}}^{(0)}\rightarrow[\widetilde{\mathbf{X}}^{(0)},\widetilde{\mathbf{X}}^{(1)},\dots,\widetilde{\mathbf{X}}^{(L)}],\widetilde{\mathbf{X}}^{(0)} = \mathbf{X}.$
            \STATE \uuline{Node-wise Message Aggregation}
            \FORALL {$i=1,2,\cdots,n$}
            \STATE Calculate weights $\mathbf{E}^{(l)} =$
            \STATE $\operatorname{MLP}\left(\operatorname{Complex}\left(\widetilde{\mathbf{X}}^{(0)}\right)\Vert\dots\Vert\left(\operatorname{Complex}\left(\widetilde{\mathbf{X}}^{(K)}\right)\right)\right)$;
            \STATE Execute aggregation $\mathbf{H} = \sum_{l=0}^K=\mathbf{W}_{node}^{(l)}\widetilde{\mathbf{X}}^{(l)},$
            \STATE $\mathbf{W}_{node}^{(l)}=e^{\delta\left(\mathbf{E}^{(l)}\right)}/\sum_{l=0}^K e^{\delta\left(\mathbf{E}^{(l)}\right)}$;
            \ENDFOR
            \STATE Calculate node soft label ${\mathbf{Z}}={\rm Softmax}\left(\mathbf{W}_{update}\mathbf{H}\right)$;
            \STATE Update trainable weights$\mathbf{W}_{update}, \mathbf{W}_{edge}, \mathbf{W}_{node}$;
            \STATE Replace the soft label of the training set node with the real label in the training sets $\mathbf{Y}_{\mathcal{V}_l}$;
            \STATE $\mathbf{Q}^{feat}=\operatorname{Norm}\left(\arccos\left(\frac{\mathbf{Z}_u\cdot\mathbf{Z}_v}{\Vert\mathbf{Z}_u \Vert\times \Vert \mathbf{Z}_v\Vert}\right)\right)$, $ \operatorname{Norm}(\mathbf{x}) = \frac{2\mathbf{x}}{\pi}$;
            \STATE Calculate node predictions or link predictions $\hat{\mathbf{Y}}$ by the soft label ${\mathbf{Z}}$ and practical downstream tasks;
        \ENDFOR
    \end{algorithmic}
\end{algorithm}

\begin{table*}[htbp]
    \caption{Algorithm complexity analysis of existing digraph neural networks.
    $n$, $m$, and $f$ are the number of nodes, edges, and feature dimensions, respectively. 
    $b$ is the batch size.
    $k$ and $K$ correspond to the $k$-order proximity of neighbors and the number of times we aggregate features. 
    $\omega$ is the time complexity of computing the approximate linear rank using Monte Carlo sampling.
    $L$ is the number of layers in learnable classifiers and $c$ represents the complex numbers consisting of real and imaginary parts.
    $H$ is the dimension of a hyperbolic space.
    $Q$ denotes the number of spectral filters.
    }
    \label{tab: algorithm_analysis}
\begin{tabular}{c|c|c|c|c|c}
\midrule[0.3pt]
Type                     & Model    & Pre-processing           & Training                 & Inference                & Memory                 \\ \midrule[0.3pt]
\multirow{2}{*}{Others}                  & D-HYPR   & $O(kKmf)$                & $O(LHKkmf+LHKknf^2)$     & $O(LHKkmf+LHKknf^2)$     & $O(bLKHf+KHf^2+kHf^2)$ \\
                         & HoloNet  & $O(m+n\log n)$           & $O(LKmf+LKnf^2+Qf^2)$         & $O(LKmf+LKnf^2+Qf^2)$         & $O(bLKf+Kf^2+n\log nf+Qf)$         \\ \midrule[0.3pt]
\multirow{6}{*}{Directed} & DGCN     & $O(m^k)$                 & $O(LKmf+LKnf^2)$         & $O(LKmf+LKnf^2)$         & $O(bLKf+Kf^2)$         \\
                         & DiGCN    & $O(km)$                  & $O(LKmf+LKnf^2)$         & $O(LKmf+LKnf^2)$         & $O(bLKf+Kf^2)$         \\
                         & NSTE     & -                        & $O(LK^kmf+LK^knf^2)$     & $O(LK^kmf+LK^knf^2)$     & $O(bLK^kf+K^kf^2)$     \\
                         & DIMPA    & $O(m)$                   & $O(LKk^2mf+LKk^2nf^2)$   & $O(LKk^2mf+LKk^2nf^2)$   & $O(bLKk^2f+k+Kf^2)$    \\
                         & Dir-GNN  & $O(km)$                  & $O(LKkmf+LKknf^2)$       & $O(LKkmf+LKknf^2)$       & $O(bLKkf+Kkf^2)$       \\
                         & ADPA     & $O(kKmf)$                & $O(kLnf^2+KLnf^2)$       & $O(kLnf^2+KLnf^2)$       & $O(bkKf+Kf^2+kf^2)$    \\ \midrule[0.3pt]
\multirow{5}{*}{Magnetic} & MagNet   & $O(m)$                   & $O(Lm^cf+Ln^cf^2)$       & $O(Lm^cf+Ln^cf^2)$       & $O(bcLf+f^2)$          \\
                         & MGC      & $O(m+\log Kcm^\omega f)$ & $O(Lnc^2f^2)$            & $O(Lnc^2f^2)$            & $O(bcLf+f^2)$          \\
                         & Framelet & $O(m+n\log n)$         & $O(Lmf+Lnf^2+Qmcf^2)$ & $O(Lmf+Lnf^2+Qmcf^2)$ & $O(bcLf+Qf+f^2+n\log nf)$       \\
                         & LightDiC & $O(m+Kcmf)$              & $O(ncf^2)$               & $O(ncf^2)$               & $O(bcKf+f^2)$          \\
                         & MAP++    & $O(n+m+Kcmf)$            & $O(mcf^2+Kncf^2)$        & $O(mcf^2+Kncf^2)$        & $O(bcKf+mcf^2+nf^2)$         \\ \midrule[0.3pt]
\end{tabular}
\end{table*}

    In this section, we provide an overview of recently proposed digraph neural networks and conduct a comprehensive analysis of their theoretical time and space complexity, as summarized in Table ~\ref{tab: algorithm_analysis}.
    To begin with, we clarify that the training and inference time complexity of the DGCN with $L$ layers and $K$ aggregators can be bounded by $O(LKmf+LKnf^2)$, where $O(LKmf)$ represents the total cost of the weight-free sparse-dense matrix multiplication in $\operatorname{Message-Agg}\left(\cdot\right)$ from Eq.~(\ref{eq: Prevalent directed Message Passing}), with DGCN utilizing GCN as the mechanism of aggregation function, and $O(LKnf^2)$ being the total cost of the feature transformation achieved by applying $K$ learnable aggregator weights.
    At first glance, $O(LKnf^2)$ may appear to be the dominant term, considering that the average degree $d$ in scale-free networks is typically much smaller than the feature dimension $f$, thus resulting in $LKnf^2 > LKndf = LKmf$.
    However, in practice, the feature transformation can be performed with significantly less cost due to the improved parallelism of dense-dense matrix multiplications.
    Consequently, $O(LKmf)$ emerges as the dominating complexity term of DGCN, and the execution of full neighbor propagation becomes the primary bottleneck for achieving scalability.

    Building upon this, we first analyze two methods (hyperbolic for D-HYPR and frequency-response filters for HoloNet), D-HYPR~\cite{zhou2022dhypr} and HoloNet~\cite{koke2023holonets}, which do not belong to the general message-passing paradigm. 
    For D-HYPR, its core lies in projecting the digraph into $H$-dimension hyperbolic space and designing $L$ trainable aggregators based on $k$-order RF and $K$-times aggregation. 
    Consequently, its time complexity can be bounded by $O(LHKkmf+LHKknf^2)$.
    As for HoloNet, it abandons the message-passing mechanism and focuses on digraph learning from a spectral perspective using holomorphic filters. 
    The key lies in Fourier transform-based spectral decomposition, with the algorithm's time complexity bounded by $O(n\log n)$. 
    Regarding the subsequent filter and corresponding learning mechanism design, it primarily depends on the size $Q$ of the filter banks, hence can be bounded by $O(Qf^2)$.
        
    Regarding methods following the prevalent directed message passing illustrated in Sec~\ref{sec: Directed Graph Neural Networks},  DiGCN~\cite{tong2020digcn} is similar to DGCN as they both use $k$-order NP as pre-processing, but the generated real symmetric adjacency matrix is different.
    DiGCN extends approximate personalized PageRank for constructing digraph Laplacian as pre-processing with time complexity of $O(m)$, which is equivalent to the undirected symmetric adjacency matrix. 
    NSTE~\cite{kollias2022nste} performs an additional aggregation based on the $k$-order proximity in each learnable aggregator, which is bounded by $O(LK^kmf+LK^knf^2)$. 
    DIMPA~\cite{he2022dimpa} extends the RF by considering incoming and outgoing edges independently in each aggregation step $O(LKk^2mf+LKk^2nf^2)$.
    Dir-GNN~\cite{dirgnn_rossi_2023} extends the $k$-order based on edge direction and encodes it using two independent sets of parameters in $L$ trainable aggregators. 
    Therefore, its time complexity can be bounded by $O(LKkmf+LKknf^2)$.
    ADPA~\cite{sun2023adpa} further employs a hierarchical attention mechanism to fuse messages for both propagation operators and receptive fields, bounded respectively by $O(Kf^2)$ and $O(kf^2)$.
    The existing methods follow directed spatial message-passing mechanisms, which inherently rely on directed edges for aggregator design, making it challenging to handle large-scale digraphs. 
    Furthermore, their use of two sets of independent learnable weights to encode source and target nodes results in a large $K$, which further exacerbates the computational costs.

    As for methods following the complex domain message passing, MAP++, MGC~\cite{zhang2021mgc}, and LightDiC~\cite{li2024lightdic} follow the decoupled paradigm, MageNet~\cite{zhang2021magnet} and Framelet~\cite{lin2023framelet_gnn} combines the propagation and training process into a deep coupled architecture. 
    In the pre-processing, all approaches achieve a time complexity of $O(m)$ to obtain the magnetic Laplacian, with the introduction of a $O(c)$ complexity due to the complex-valued matrix.  
    Then, MGC conducts multiple graph propagation approximately with significantly larger $K$, bounded by $O(\log Kcm^\omega f)$.
    Framelet employs a spectral decomposition similar to HoloNet. 
    However, Framelet extends the concept of wavelet transforms by integrating short-duration signals from different frequency bands to achieve more comprehensive data processing in signal representation.
    In contrast, MAP++ and LightDiC perform only a finite number of graph propagation with small $K$, bounded by $(OKcmf)$.
    In the training, as the magnetic Laplacian involves real and imaginary parts, the fully square recursive computation cost of MagNet and Framelet grows exponentially with the increase of the number of nodes and edges, reaching $O(Lm^cf+Ln^cf^2)$ and $O(Qmcf^2)$. 
    In contrast, MGC performs complex-valued forward propagation with a complexity of $(Lnc^2f^2)$, while LightDiC further decouples the complex-valued matrices and reduces the computation complexity to $O(ncf^2)$ by employing the simple linear logistic regression.
    Although their neural architectures are simple, they often encounter performance limitations when dealing with complex digraphs.
    Therefore, in MAP++, we introduce edge-wise graph propagation and node-wise message aggregation.
    Notably, the former operates only on directed structural entropy and local clustering coefficients, resulting in negligible computational overhead. 
    Meanwhile, the computational complexity of the latter is strictly bounded by $O(Kncf^2)$. 
    Furthermore, during iterative training, we can intentionally reduce the encoding frequency to further reduce overhead.

\subsection{The Proof of Theorem~\ref{theorem: 1}}
\label{appendix:The Proof of Theorem1}
\begin{proof}
    To prove the skew-symmetry of $\boldsymbol\Theta^{\left(q^\star\right)}$ and the Hermitian property of $\exp \left(i \boldsymbol\Theta^{\left(q^\star\right)}\right)$, we begin by analyzing the relationships established in Eq.~(\ref{eq: q topology encoding})-(\ref{eq: map framework}).
    
    From Eq.~(\ref{eq: q topology encoding}), we observe that $q_{u v}^{\operatorname {topo }}=q_{v u}^{\operatorname {topo }}$, indicating that the topological contribution to the parameter $q$ between nodes $u$ and $v$ is symmetric. 
    Similarly, from Eq.~(\ref{eq: q feature encoding}), we find that $q_{u v}^{\operatorname {feat }}=$ ${q}_{v u}^{\operatorname {feat }}$, confirming that the feature-based contribution to ${q}$ is also symmetric. 
    Therefore, by combining these two components in Eq.~(\ref{eq: map framework}), we conclude that $q_{u v}^{\star}=q_{v u^{\prime}}^{\star}$ meaning that the overall parameter $q^{\star}$ is symmetric with respect to nodes $u$ and $v$.
    
    Next, using this symmetry, we examine the matrix $\boldsymbol\Theta^{\left(q^\star\right)}$, which encodes the phase differences between nodes in the complex domain. 
    Specifically, we have:$\boldsymbol\Theta^{\left(q^\star\right)}(u, v)=-\boldsymbol\Theta^{\left(q^\star\right)}(v, u)$.
    This relationship confirms that $\boldsymbol\Theta^{\left(q^{\star}\right)}$ is skew-symmetric, meaning that $\boldsymbol\Theta^{\left(q^{\star}\right)}=-\left(\boldsymbol\Theta^{\left(q^{\star}\right)}\right)^{\top}$. 
    Here, for any real skew-symmetric matrix $\mathbf{A}$, the matrix $\exp (i \mathbf{A})$ is Hermitian.
    $\boldsymbol\Theta^{\left(q^{\star}\right)}$ is skew-symmetric, it follows that $\exp \left(i \boldsymbol\Theta^{\left(q^\star\right)}\right)$ is a Hermitian matrix. 
    This property is crucial in ensuring that the matrix captures the directed dependencies between nodes in a way that preserves the necessary mathematical structure for subsequent analysis.
    In summary, the symmetry of $q^{\star}$ leads to the skew-symmetry of $\boldsymbol\Theta^{\left(q^{\star}\right)}$, and as a result, $\exp \left(i \boldsymbol\Theta^{\left(q^\star\right)}\right)$ is Hermitian, confirming the desired properties.
\end{proof}

\subsection{The Proof of Theorem~\ref{theorem: 2}}
\label{appendix:The Proof of Theorem2}
\begin{proof}
    We suppose the proportion of noisy offsets in $\boldsymbol{\Theta}$ is $1-p$. 
    Let $z$ be the normalized vector defined as $z_i=\frac{e^{iw_i}}{\sqrt{n}}$. 
    With a probability of $p$, the edge $\{i,j\}$ is good and $\mathbf{H}_{ij}=e^{i(w_i-w_j)}$. 
    On the other hand, with a probability of $1-p$, the edge is bad. 
    The matrix $\mathbf{H}$ can be decomposed as $\mathbf{H}=npzz^*+\mathbf{R}$, where $\mathbf{R}$ is a noise matrix.

    According to~\cite{singer2011graph_angular_synchronization1}, the correlation between $v_1$ and $z$ can be predicted by using regular perturbation theory for solving the eigenvector equation in an asymptotic expansion.
    In quantum mechanics~\cite{griffiths2018graph_angular_synchronization6}, the asymptotic expansions of the non-normalized eigenvector $v_1$ is given by
\begin{equation}
    v_1 \sim z+\frac{\mathbf{R}z-(z^*\mathbf{R}z)z}{np}+\dots.
\end{equation}
Because $||\mathbf{R}z-(z^*\mathbf{R}z)z||^2=||\mathbf{R}z||^2-(z^*\mathbf{R}z)^2$, the angle $\alpha$ between the eigenvector $v1$ and the vector of true attributes $z$ satisfies the asymptotic relation
\begin{equation}
\label{angle}
    \tan^2\alpha \sim \frac{||\mathbf{R}z||^2-(z^*\mathbf{R}z)^2}{(np)^2}+\dots.
\end{equation}
    The expected values of the numerator terms in Eq.~(\ref{angle}) are given by
\begin{equation}
\begin{aligned}
\mathbb{E}||\mathbf{R}z||^2
&=\mathbb{E}\sum^n_{i=1}\left|\sum^n_{j=1}\mathbf{R}_{ij}z_j\right|^2 \\
&=\sum^n_{i,j=1}{\rm Var}(\mathbf{R}_{ij}z_j) \\
&=\sum^n_{i=1}\sum_{j\neq i}|z_j|^2(1-p^2) \\
&=(n-1)(1-p^2) ,
\end{aligned}
\end{equation}
and
\begin{equation}
\begin{aligned}
\mathbb{E}(z^*\mathbf{R}z)^2
&=\mathbb{E}\left[\sum^n_{i,j=1}\mathbf{R}_{ij}\overline{z_i}z_j \right]^2 \\
&=\sum^n_{i,j=1}{\rm Var}(\mathbf{R}_{ij}\overline{z_i}z_j) \\
&=(1-p^2)\sum_{i\neq j}|z_i|^2|z_j|^2 \\
&=(1-p^2)\left[\left(\sum^n_{i=1}|z_i|^2\right)^2-\sum^n_{i=1}|z_i|^4 \right] \\
&=(1-p^2)\left(1-\frac{1}{n}\right),
\end{aligned}
\end{equation}
because the variance of $\mathbf{R}$ is given by $1-p^2$~\cite{singer2011graph_angular_synchronization1} and $|z_i|^2=\frac{1}{n}$. 
    Based on the above three formulas, we can derive the following:
\begin{equation}
\mathbb{E}\tan^2\alpha\sim\frac{(n-1)^2(1-p^2)}{n^3p^2}+\dots.
\end{equation}
In the vast majority of cases, $n\gg 1$ and $p \ll 1$, thus we further obtain the following formula:
\begin{equation}
\mathbb{E}\tan^2\alpha\sim\frac{1}{np^2}+\dots.
\end{equation}
    This formulation indicates that when $np^2$ approaches infinity, the angle between $v_1$ and $z$ tends to zero and the correlation between them approaches $1$. 
    We further infer that even for extremely small $p$ values, the eigenvector method effectively retrieves the attributes if there are sufficient equations, meaning if $np^2$ is adequately large.
\end{proof}

\subsection{The Proof of Theorem~\ref{theorem: 3}}
\label{appendix:The Proof of Theorem3}
\begin{proof}
    Since $\boldsymbol\Theta$ is noise-free, we define $\mathbf{K}$ as an $n\times n$ matrix where $\mathbf{K}_{ij}=1$ for all $i,j$. 
    As the $\mathbf{K}$ is symmetric, it possesses a complete set of real eigenvalues $\lambda_1\geq\lambda_2\geq\dots\geq\lambda_n$, along with corresponding real orthonormal eigenvectors $\psi_1,\cdots,\psi_n$.
    We can express $\mathbf{K}$ in terms of its eigenvalues and eigenvectors as follows:
\begin{equation}
\mathbf{K}=\sum_{l=1}^n\lambda_l\psi_l\psi_l^T.
\end{equation}
    Next, let $\mathbf{Z}$ be an $n\times n$ diagonal matrix with diagonal elements $\mathbf{Z}_{ii}=e^{iw_i}$. 
    It is evident that $\mathbf{Z}$ is a unitary matrix, satisfying $\mathbf{Z}\mathbf{Z}^*=\mathbf{I}$. 
    We then construct the Hermitian matrix $\mathbf{B}$ by conjugating $\mathbf{K}$ with $\mathbf{Z}$:
\begin{equation}
\mathbf{B}=\mathbf{Z}\mathbf{K}\mathbf{Z}^*.
\end{equation}
    The eigenvalues of $\mathbf{B}$ remain the same as those of $\mathbf{K}$, namely $\lambda_1,\lambda_2,\dots,\lambda_n$. 
    The corresponding eigenvectors $\{\phi_l\}^n_{l=1}$ of $\mathbf{B}$, satisfying $\mathbf{B}\phi_l=\lambda_l\phi_l$, are given by
\begin{equation}
\phi_l=\mathbf{Z}\psi_l,\;\;\;l=1,\dots,n.
\end{equation}
    Next, we observe the entries of $\mathbf{B}$:
\begin{equation}
\mathbf{B}_{ij}=e^{i(w_i-w_j)}.
\end{equation}
    According to the Perron-Frobenius theorem~\cite{horn2012graph_angular_synchronization5}, since $\mathbf{K}$ is a non-negative matrix, the components of the top eigenvector $\psi_1$ associated with the largest eigenvalue $\lambda_1$ are all positive:
\begin{equation}
\psi_1(i)>0,\;\;\forall i=1,2,\dots,n.
\end{equation}
    Consequently, we examine the complex phases of the coordinates of the top eigenvector $\phi_1=\mathbf{Z}\psi_1$. 
    Thus, the complex phases of the coordinates of $\phi_1$ are identical to the true attributes:
\begin{equation}
e^{i\hat{w}_i}=\frac{\phi_1(i)}{|\phi_1(i)|}.
\end{equation}
\end{proof}

\subsection{Our Approach and GNNSync}
\label{appendix: Our Approach and GNNSync}
    The attribute synchronization problem we propose can also be addressed by GNNSync~\cite{he2023graph_angular_synchronization4}. 
    It reframes the synchronization problem as a theoretically grounded digraph learning task, where angles are estimated by designing a specific GNN architecture to extract graph embeddings and leveraging newly introduced loss functions. 
    This method has demonstrated superior performance in high-noise environments.
    Notably, our proposed MAP framework can further enhance the attribute synchronization process when integrated with GNNSync in the following two significant ways.
    
    Firstly, MAP can act as an encoder within the GNNSync framework, generating higher-quality node embeddings compared to DIMPA used in the original implementation. 
    By more effectively encoding both node features and topology, MAP improves the overall learning capability of the model.
    Secondly, the adaptive phase matrix introduced by MAP enables personalized encoding of directed edges, capturing critical directed information. 
    This personalized encoding allows the generated node attributes to more accurately reflect the underlying characteristics of each node, ultimately improving the performance of the synchronization task. 
    Through these enhancements, the MAP framework positions itself as a powerful tool for advancing the capabilities of GNNSync and other similar methods in digraph learning and attribute synchronization.

\subsection{Dataset Description}
\label{appendix: Dataset Description}

\begin{table*}[htbp]
\setlength{\abovecaptionskip}{0.2cm}
\setlength{\belowcaptionskip}{-0.2cm}
\caption{The statistical information of the experimental datasets.
}
\label{tab: datasets}
\begin{tabular}{ccccccc}
\midrule[0.3pt]
Datasets        & \#Nodes     & \#Edges       & \#Features & \#Classes & \#Train/Val/Test & Description            \\ \midrule[0.3pt]
CoraML          & 2,995       & 8,416         & 2,879      & 7         & 140/500/2,355    & citation network       \\
CiteSeer        & 3,312       & 4,591         & 3,703      & 6         & 120/500/2,692    & citation network       \\
Actor           & 7,600       & 26,659        & 932        & 5         & 48\%/32\%/20\%   & actor network          \\
WikiCS          & 11,701      & 290,519       & 300        & 10        & 580/1,769/5,847  & weblink network        \\
Tolokers        & 11,758      & 519,000       & 10         & 2         & 50\%/25\%/25\%   & crowd-sourcing network \\
Empire          & 22,662      & 32,927        & 300        & 18        & 50\%/25\%/25\%   & article syntax network \\
Rating          & 24,492      & 93,050        & 300        & 5         & 50\%/25\%/25\%   & rating network         \\
ogbn-arXiv      & 169,343     & 2,315,598     & 128        & 40        & 91k/30k/48k      & citation network       \\
ogbn-papers100M & 111,059,956 & 1,615,685,872 & 128        & 172       & 1207k/125k/214k  & citation network       \\\midrule[0.3pt]
Slashdot        & 75,144      & 425,702     & 100        & Link-level       & 80\%/15\%/5\%  & social network       \\
Epinions & 114,467 & 717,129 & 100        & Link-level       & 80\%/15\%/5\%  & social network       \\
WikiTalk & 2,388,953 & 5,018,445 & 100        & Link-level       & 80\%/15\%/5\%  & co-editor network       \\
\midrule[0.3pt]
\end{tabular}
\end{table*}

    In our experiments, we evaluate the performance of our proposed MAP and MAP++ on 12 digraph benchmark datasets. 
    The 12 publicly available digraph datasets are sourced from multiple domains, highlighting the comprehensive nature of our experiments. 
    Specifically, they include 4 citation networks (CoraML, Citeseer, ogbn-arXiv, and ogbn-papers100M) in~\cite{bojchevski2018coraml_citeseer,mernyei2020wikics, hu2020ogb}, actor network (Actor)~\cite{pei2020geomgcn}, web-link network (WikiCS) in~\cite{mernyei2020wikics}, crowd-sourcing network (Toloklers)~\cite{platonov2023hete_gnn_survey4}, e-commerce network (Rating)~\cite{platonov2023hete_gnn_survey4}, syntax network (Empire)~\cite{platonov2023hete_gnn_survey4}, 2 social networks (Slashdot and Epinions) in~\cite{ordozgoiti2020slashdot,massa2005epinions}, and co-editor network~\cite{leskovec2010wikitalk}.
    The dataset statistics are shown in Table~\ref{tab: datasets} and more descriptions can be found later.
    
    Notably, given MAP and MAP++ focus on providing tailored solutions for complex domain message passing based on the magnetic Laplacian, and considering that directed information is disregarded in undirected graphs, we opted not to use undirected graphs as validation datasets and instead focused our efforts on digraph benchmark datasets.
    
    We need to clarify that we are using the directed version of the dataset instead of the one provided by the PyG library (CoraML, CiteSeer)\footnote{https://pytorch-geometric.readthedocs.io/en/latest/modules/datasets.html}, WikiCS paper\footnote{https://github.com/pmernyei/wiki-cs-dataset} and the raw data given by the OGB (ogb-arxiv)\footnote{https://ogb.stanford.edu/docs/nodeprop/}.
    Meanwhile, we remove the redundant multiple and self-loop edges to further normalize the 10 digraph datasets.
    In addition, for Slashdot, Epinions, and WikiTalk, the PyGSD~\cite{he2023pygsd} library reveals only the topology and lacks the corresponding node features and labels.
    Therefore, we generate the node features using eigenvectors of the regularised topology.
    Building upon this foundation, the description of all digraph benchmark datasets is listed below:

    \textbf{CoraML} and \textbf{CiteSeer}~\cite{bojchevski2018coraml_citeseer} are two citation network datasets. 
    In these two networks, papers from different topics are considered nodes, and the edges are citations among the papers. 
    The node attributes are binary word vectors, and class labels are the topics the papers belong to.

    \textbf{Actor}~\cite{pei2020geomgcn} is an actor co-occurrence network in which nodes denote actors, and edges signify actors appearing together on Wikipedia pages. 
    Node features are bag-of-words vectors derived from keywords found on these Wikipedia pages. 
    They are categorized into five groups based on the terms found in the respective actor's Wikipedia page.

    \textbf{WikiCS}~\cite{mernyei2020wikics} is a Wikipedia-based dataset for bench-marking GNNs. 
    The dataset consists of nodes corresponding to computer science articles, with edges based on hyperlinks and 10 classes representing different branches of the field.
    The node features are derived from the text of the corresponding articles. 
    They were calculated as the average of pre-trained GloVe word embeddings~\cite{pennington2014glove}, resulting in 300-dimensional node features.

    \textbf{Tolokers}~\cite{platonov2023hete_gnn_survey4} is derived from the Toloka crowdsourcing platform~\cite{Tolokers_original}.
    Nodes correspond to tolokers (workers) who have engaged in at least one of the 13 selected projects. 
    An edge connects two tolokers if they have collaborated on the same task. 
    The objective is to predict which tolokers have been banned in one of the projects.
    Node features are derived from the worker's profile information and task performance statistics.

    \textbf{Empire}~\cite{platonov2023hete_gnn_survey4} is based on the Roman Empire article from the English Wikipedia~\cite{lhoest2021empire_original}, each node in the graph corresponds to a non-unique word in the text, mirroring the article's length. 
    Nodes are connected by an edge if the words either follow each other in the text or are linked in the sentence's dependency tree. 
    Thus, the graph represents a chain graph with additional connections.

    \textbf{Rating}~\cite{platonov2023hete_gnn_survey4} is derived from the Amazon co-purchasing network metadata available in the SNAP\footnote{https://snap.stanford.edu/}~\cite{leskovec2014rating_original}. 
    Nodes are products, and edges connect items bought together. 
    The task involves predicting the average rating given by reviewers, categorized into five classes. 
    Node features are based on the mean FastText embeddings~\cite{grave2018fast_word_embedding} of words in the product description.
    To manage graph size, only the largest connected component of the 5-core is considered.

    \textbf{Ogbn-arxiv} and \textbf{ogbn-papers100M}~\cite{hu2020ogb} are two citation graphs indexed by MAG~\cite{wang2020microsoft_MAG}.
    For each paper, we generate embeddings by averaging the word embeddings from both its title and abstract. 
    These word embeddings are computed using the skip-gram model, which captures the semantic relationships between words based on their context. 
    This approach allows us to create a comprehensive representation of the paper's content.

    \textbf{Slashdot}~\cite{ordozgoiti2020slashdot} is from a technology-related news website with user communities. 
    The website introduced Slashdot Zoo features that allow users to tag each other as friends or foes. 
    The dataset is a common signed social network with friends and enemies labels.
    In our experiments, we only consider friendships.
    
    \textbf{Epinions}~\cite{massa2005epinions} is an online social network centered around "who-trusts-whom" dynamics relationship systems, where users can indicate trust or distrust tags in the reviews and opinions uploaded by other users.
    This network captures social interactions and the formation of trust within the community. 
    For the purposes of our experiments, we focus solely on the "trust" relationships, excluding the "distrust" connections to streamline our analysis.

    \textbf{WikiTalk}~\cite{leskovec2010wikitalk} includes all users and discussions from the inception of Wikipedia until January 2008. 
    The network comprises $n=2,388,953$ nodes, where each node represents a Wikipedia user, and a directed edge from node $v_i$ to node $v_j$ indicates that user $i$ edited user $j$ 's talk page at least once. 
    For our analysis, we extract the largest weakly connected component.
    
\subsection{Compared Baselines}
\label{appendix: Compared Baselines}

    The baselines we employ are as follows:
    (1) Directed prevalent message passing-based approaches: DGCN~\cite{tong2020dgcn}, DiGCN~\cite{tong2020digcn}, DIMPA~\cite{he2022dimpa}, NSTE~\cite{kollias2022nste}, Dir-GNN~\cite{dirgnn_rossi_2023}, and ADPA~\cite{sun2023adpa}; 
    (2) Directed MagDGs: MagNet~\cite{zhang2021magnet}, MGC\cite{zhang2021mgc}, Framelet-Mag (Framelet)~\cite{lin2023framelet_gnn}, LightDiC~\cite{li2024lightdic}.
    (3) Undirected methods and other digraph neural networks: GCN~\cite{kipf2016gcn}, GAT~\cite{velivckovic2017gat}, GCNII~\cite{chen2020gcnii}, GATv2~\cite{brody2021gatv2}, OptBasisGNN~\cite{OptBasisGNN} (OptBG), NAGphormer~\cite{chen2022nagphormer} (NAG), GAMLP~\cite{gamlp}, D-HYPR~\cite{zhou2022dhypr}, and HoloNet~\cite{koke2023holonets}.
    Notably, to verify the generalization of our proposed MAP and MAP++, we compare the undirected GNNs in digraphs with coarse undirected transformation (i.e., convert directed edges into undirected edges).
    The descriptions of them can be found later in this section.
    To alleviate the influence of randomness, we repeat each experiment 10 times to represent unbiased performance and running time (second report).
    Meanwhile, we present experiment results with various baselines in separate modules, avoiding abundant charts and validating the generalizability of our proposed methods.

    \textbf{DGCN}~\cite{tong2020dgcn}: DGCN proposes the first and second-order proximity of neighbors to design a new message-passing mechanism, which in turn learns aggregators based on incoming and outgoing edges using two sets of independent learnable parameters.

    \textbf{DiGCN}~\cite{tong2020digcn}: DiGCN notices the inherent connections between graph Laplacian and stationary distributions of PageRank, it theoretically extends personalized PageRank to construct real symmetric Digraph Laplacian. Meanwhile, DiGCN uses first-order and second-order neighbor proximity to further increase RF.
    
    \textbf{DIMPA}~\cite{he2022dimpa}: DIMPA represents source and target nodes separately.
    It performs a weighted average of the multi-hop neighborhood information to capture the local network information.
    
    \textbf{NSTE}~\cite{kollias2022nste}: NSTE is inspired by the 1-WL graph isomorphism test, which uses two sets of trainable weights to encode source and target nodes separately. Then, the information aggregation weights are tuned based on the parameterized feature propagation process.

    \textbf{Dir-GNN}~\cite{dirgnn_rossi_2023}: Dir-GNN introduces a versatile framework tailored for heterophilous settings. 
    It addresses edge directionality by conducting separate aggregations of incoming and outgoing edges. 
    Demonstrated to match the expressivity of the directed Weisfeiler-Lehman test, Dir-GNN outperforms conventional MPNNs in accurately modeling digraphs.

    \textbf{ADPA}~\cite{sun2023adpa}: ADPA adaptively explores suitable directed k-order neighborhood operators to conduct weight-free graph propagation and employs two hierarchical node-adaptive attention mechanisms to acquire optimal node representations.

    \textbf{MagNet}~\cite{zhang2021magnet}: MagNet utilizes complex numbers to model directed information, it proposes a spectral GNN for digraphs based on a complex Hermitian matrix known as the magnetic Laplacian. Meanwhile, MagNet uses additional trainable parameters to combine the real and imaginary filter signals separately to achieve better prediction performance.
    
    \textbf{MGC}~\cite{zhang2021mgc}: MGC introduces the magnetic Laplacian, a discrete operator with the magnetic field, which preserves edge directionality by encoding it into a complex phase with an electric charge parameter. 
    By adopting a truncated variant of PageRank named Linear-Rank, it designs and builds a low-pass filter for homogeneous graphs and a high-pass filter for heterogeneous graphs.

    \textbf{Framelet}~\cite{lin2023framelet_gnn}: Framelet utilizes the framelet transform to enhance the representation of digraph signals. 
    These framelets are constructed using the complex-valued magnetic Laplacian, enabling signal processing in both real and complex domains simultaneously.

    \textbf{LightDiC}~\cite{li2024lightdic}:  LightDiC is a scalable adaptation of digraph convolution built upon the magnetic Laplacian, which performs topology-related computations during offline pre-processing.

    \textbf{GCN}~\cite{kipf2016gcn}: GCN is guided by a localized first-order approximation of spectral graph convolutions. This model's scalability is directly proportional to the number of edges, and it learns intermediate representations in hidden layers that capture both the topology and node features.

    \textbf{GCNII}~\cite{chen2020gcnii}: GCNII incorporates initial residual and identity mapping. Theoretical and empirical evidence is presented to demonstrate how these techniques alleviate the over-smoothing problem.
    
    \textbf{GAT}~\cite{velivckovic2017gat}: GAT utilizes attention mechanisms to quantify the importance of neighbors for message aggregation.
    This strategy enables implicitly specifying different weights to different nodes in a neighborhood, without depending on the graph structure upfront.

    \textbf{GATv2}~\cite{brody2021gatv2}: GATv2 introduces a variant with dynamic graph attention mechanisms to improve GAT.
    This strategy provides better node representation capabilities and enhanced robustness when dealing with graph structure as well as node attribute noise.

    \textbf{OptBasisGNN}~\cite{OptBasisGNN}: OptBasisGNN revolutionizes GNNs by redefining polynomial filters. 
    It dynamically learns suitable polynomial bases from training data, addressing fundamental adaptability.

    \textbf{NAGphormer}~\cite{chen2022nagphormer} treats each node as a sequence containing a series of tokens. 
    For each node, it aggregates the neighborhood features from different hops into different representations.

    \textbf{GAMLP}~\cite{gamlp}: GAMLP is designed to capture the inherent correlations between different scales of graph knowledge to break the limitations of the enormous size and high sparsity level of graphs hinder their applications under industrial scenarios.
    
    \textbf{D-HYPR}~\cite{zhou2022dhypr}: D-HYPR introduces hyperbolic from diverse neighborhoods. 
    This conceptually simple yet effective framework extends seamlessly to digraphs with cycles and non-transitive relations, showcasing versatility in various downstream tasks.

    \textbf{HoloNet}~\cite{koke2023holonets}: HoloNet demonstrates that spectral convolution can extend to digraphs. By leveraging advanced tools from complex analysis and spectral theory, HoloNet introduces spectral convolutions tailored for digraphs.
    
\subsection{Hyperparameter Settings}
\label{appendix: Hyperparameter Settings}
    The hyperparameters in the baseline models are set according to the original paper if available.
    Otherwise, we perform a hyperparameter search via the Optuna~\cite{akiba2019optuna}.
    For both our proposed methods, MAP and MAP++, their satisfactory flexibility in method and neural architecture design obviates the need for additional hyperparameter search. 
    However, we recommend exploring the number of graph propagation steps and the dimension of hidden embeddings within the range of $[3, 10]$ and $[64, 128, 256, 512]$ to further enhance predictive performance.
    Regarding the experimental results of Dir-GNN and HoloNet on the Empire dataset, we would like to clarify that we ensured a fair comparison by using a class-balanced dataset split instead of the pre-split datasets used in Dir-GNN and HoloNet.
    
\subsection{Experiment Environment}
\label{appendix: Experiment Environment}
    The experiments are conducted on the machine with Intel(R) Xeon(R) Gold 6240 CPU @ 2.60GHz, and NVIDIA A100 80GB PCIe and CUDA 12.2. The operating system is Ubuntu 18.04.6 with 216GB memory.
    As for software versions in the environment, we use Python 3.9 and Pytorch 1.11.0.
    
\subsection{$q$ Selection in Spectral Graph Theory}
\label{appendix: Guidance for Selecting q in Spectral Graph Theory}

(1) Directed Edges Num~\cite{geisler2023transformers_meet_digraph}:
    It posits that the potential $q$ governs the magnitude of the induced phase shift by each edge. 
    Specifically, in its application to digraph-level classification and regression, $q$ assumes a role akin to the lowest frequency in sinusoidal positional encodings (typically $1 /(2 \pi \times 10,000))$. 
    Following the sinusoidal encoding convention, one could fix $q$ to a suitable value for the largest expected graphs. 
    However, in their experiments, they found that scaling the potential $q$ with the number of nodes $n$ and the quantity of directed edges leads to marginally better performance.
    In other words, they suggest opt for $q=q^{\prime} / d_{\mathrm{G}}$ with $q^{\prime}$ as the relative potential and $d_{\mathrm{G}}$ as the graph-specific normalizer. 
    This normalizer is an upper bound on the number of directed edges in a simple path, calculated as $d_{\mathrm{G}}=\max (\min (\vec{m}, n), 1)$, where $\vec{m}$ denotes the count of purely directed edges ( $(u, v) \in E$ where $(v, u) \notin E)$.

(2) Digraph Ring Length~\cite{fanuel2018magLaplacian1, fanuel2017q_magnetic3}:
    It proposes a unique perspective on the $q$, suggesting its suitability for positional encodings. 
    Specifically, in graph visualization, the selection of $q$ is related to the ring length. 
    If the ring length is $k$, then $q=1 / k$. 
    They advocate for selecting $q$ as a rational number, such as $q=1 / 3$, which proves effective for visualizing graphs comprising directed triangles. 
    In this context, each edge within a directed triangle induces a $2 / 3 \pi$ shift, resulting in a cumulative shift of 360 degrees for the triangle.

(3) Eigenvector Perturbation~\cite{furutani2020magLaplacian2}:
    It claims that the choice of the rotation parameter $q$ influences the graph Fourier transform. 
    They propose an expedient method to select $q$ for graph signal processing. 
    Let $\epsilon$ be the tolerance of the smallest eigenvalue $\lambda_0^{\prime}$ of the Hermitian Laplacian $\mathcal{L}_m^{(q)}(q>0)$ of an unweighted directed graph $\mathcal{G}$, that is $0 \leq \lambda_0^{\prime} \leq \epsilon$. 
    Then, they denote the eigenvalue and associated eigenvector of the symmetrized Laplacian $\boldsymbol{L}^{(s)}\left(=\mathcal{L}_0\right)$ of $\mathcal{G}^{(s)}$ as $\lambda_\mu^{(s)}$ and $\boldsymbol{u}_\mu^{(s)}$, respectively. 
    According to eigenvalue perturbation theory~\cite{ngo2005eigenvalue_perturbation_theory}, they obtain $0 \leq q \leq \frac{\cos ^{-1}(1-2 \epsilon /\langle d\rangle)}{2 \pi}$ .
    Thus, one can choose $q$ depending only on the average degree $\langle d\rangle$ and the tolerance $\epsilon$ of the smallest eigenvalue $\lambda_0^{\prime}$.

\end{document}